\documentclass[10pt,twocolumn,letterpaper]{article}

\usepackage[pagenumbers]{cvpr} 

\usepackage{multirow}
\usepackage[table,xcdraw]{xcolor}
\usepackage{makecell}

\definecolor{cvprblue}{rgb}{0.21,0.49,0.74}
\usepackage[pagebackref,breaklinks,colorlinks,allcolors=cvprblue]{hyperref}

\definecolor{chrxcolor}{RGB}{1, 140, 116}

\def\confName{CVPR}
\def\confYear{2027}

\title{Beyond Static Semantics: Instance-Adaptive Vision-Language Alignment for Building Damage Assessment}

\author{
Hang Yin$^{1}$ \quad
Jiepan Li$^{2,*}$ \quad
Hongruixuan Chen$^{3}$ \quad
Guangyi Yang$^{1}$ \quad
Wei He$^{2}$\\[4pt]
$^{1}$School of Electronic Information, Wuhan University\\
$^{2}$LIESMARS, Wuhan University\\
$^{3}$The University of Tokyo\\[4pt]
{\small
\texttt{\{whuyh2004, jiepanli, ygy, weihe1990\}@whu.edu.cn}
\quad
\texttt{Qschrx@gmail.com}}\\[2pt]
{\small $^{*}$Corresponding author}
}
\begin{document}
\maketitle

\begin{abstract}
Instance-level building damage assessment (IBDA) from bi-temporal remote sensing imagery remains challenging, as the same damage grade may exhibit diverse visual manifestations across individual buildings. Natural language offers explicit descriptions of physical damage, providing complementary semantic cues beyond discrete category labels. However, existing vision--language approaches have not fully addressed how damage semantics should adapt to local bi-temporal evidence and subsequently serve as instance-specific references for severity discrimination. To bridge this gap, we propose LAMIA-IBDA, a unified vision--language framework that progressively grounds shared damage semantics in local observations and individual buildings. Specifically, Adaptive Text-Visual Alignment (ATVA) enables evidence-adaptive semantic interaction by jointly contextualizing multiple damage descriptions and bi-temporal visual features under reliability-aware attention supervision. Dynamic Instance-Prototype Grounding (DIPG) further associates these contextualized semantics with individual building regions to construct instance-adaptive prototypes for fine-grained damage classification. Instance-level description preferences additionally supervise fusion-stage attention, linking semantic adaptation with instance discrimination. Extensive experiments on xBD and BRIGHT demonstrate state-of-the-art performance across optical--optical and optical--SAR settings. The source code is publicly available \href{https://github.com/Cutensil/LAMIA-IBDA}{here}.
\end{abstract}

\section{Introduction}
\label{sec:intro}

Building damage assessment (BDA) from pre- and post-disaster remote sensing (RS) imagery is critical for rapid disaster response and recovery~\cite{gupta2019xbd,zheng2021changeos}. Existing approaches have evolved from pixel-level damage mapping~\cite{daudt2018fully,chen2020stanet,fang2021snunet,chen2022bit,bandara2022changeformer,chen2024changemamba} toward instance-level BDA (IBDA), which explicitly delineates individual buildings and assigns damage severity to each instance~\cite{ma2024maskview,chen2025bright,chen2026brightchallenge}. By preserving individual building identities, instance-level modeling alleviates fragmented predictions and inconsistent building-wise labels. However, accurate instance delineation does not necessarily guarantee reliable damage assessment. Identifying individual buildings and interpreting their damage severity involve fundamentally different challenges: \textbf{\textit{the former relies on spatial localization, whereas the latter requires distinguishing the physical damage characteristics underlying different severity levels}}.

\begin{figure}[t]
\centering
\includegraphics[width=\linewidth]{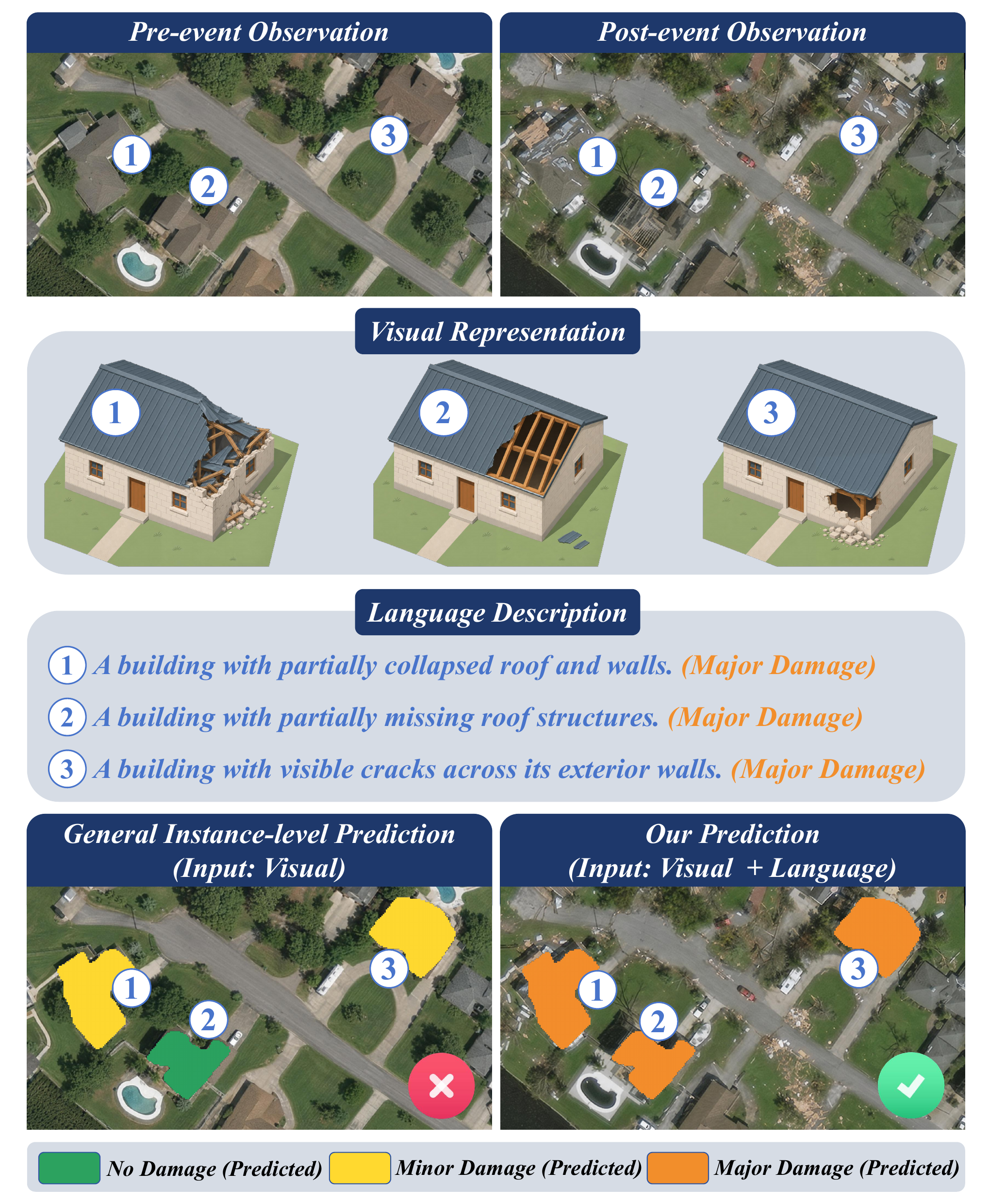}
\caption{Illustration of diverse visual manifestations within a shared damage grade and the benefit of language-guided instance-level assessment. Three buildings exhibit distinct structural damage patterns but share the same major-damage grade. Visual-only predictions yield inconsistent severity estimates, whereas language-guided predictions better identify their common damage category.}
\label{fig:motivation}
\end{figure}

A central challenge in IBDA lies in the heterogeneous visual manifestations of shared damage semantics. Buildings assigned to the same damage grade may exhibit fundamentally different physical destruction patterns, such as roof damage, structural deformation, or partial collapse, while individual buildings within the same disaster scene may present distinct damage evidence. As illustrated in Fig.~\ref{fig:motivation}, three buildings exhibit different structural damage patterns but share the same major-damage grade. This highlights the non-unique correspondence between visual evidence and damage severity, which cannot be adequately captured by learning visual representations under discrete category supervision alone~\cite{ma2024maskview,chen2025bright}. More fundamentally, \textbf{\textit{damage grades define the severity of destruction, whereas visual observations reveal how that destruction manifests in individual buildings}}. Natural language provides an explicit means of characterizing these diverse physical manifestations through complementary damage descriptions, offering semantic references beyond discrete category labels. However, the relevance of these descriptions varies across buildings and depends on their local bi-temporal evidence. Effective IBDA therefore requires damage semantics to be dynamically adapted to individual visual observations, rather than represented by fixed category embeddings.

Vision--language learning has demonstrated substantial potential in connecting visual representations with semantic knowledge~\cite{radford2021clip,rao2022denseclip,li2022lseg,cho2024catseg,liu2024remoteclip}. Existing approaches primarily leverage language through two complementary paradigms: incorporating textual semantics into visual feature learning, as exemplified by LAVT~\cite{yang2022lavt} and MMChange~\cite{zhou2025mmchange}, or aligning visual regions with language representations for semantic prediction, as in GLIP~\cite{li2022grounded}. Related studies have further explored language-guided RS change understanding and building damage reasoning~\cite{dong2024changeclip,varghese2025viewdelta,wang2026bdachat}. Despite these advances, two key challenges remain insufficiently addressed in language-guided IBDA: \textbf{\textit{(i) evidence-adaptive semantic interaction}}, which requires multiple descriptions of the same damage grade to adapt to local bi-temporal evidence while accounting for the varying reliability of visual observations; and \textbf{\textit{(ii) instance-specific semantic grounding}}, which requires locally contextualized damage semantics to be further associated with individual buildings, providing distinct classification references according to their respective damage manifestations. Addressing these challenges requires connecting local vision--language interaction with instance-level semantic discrimination, rather than treating feature fusion and region--text matching as independent operations.

To address these challenges, we propose \textbf{LAMIA-IBDA} (\textbf{L}anguage-guided \textbf{A}daptive \textbf{M}ultimodal \textbf{I}ntegration and \textbf{A}lignment for \textbf{I}nstance-level \textbf{B}uilding \textbf{D}amage \textbf{A}ssessment), a unified vision--language framework that progressively adapts shared damage semantics to local bi-temporal evidence and individual buildings. To achieve \emph{evidence-adaptive semantic interaction}, we introduce \textbf{Adaptive Text-Visual Alignment} (ATVA), which jointly contextualizes bi-temporal visual features and multiple damage descriptions under reliability-aware attention supervision. To enable \emph{instance-specific semantic grounding}, we develop \textbf{Dynamic Instance-Prototype Grounding} (DIPG), which establishes spatial correspondence between contextualized descriptions and candidate building regions, constructing instance-adaptive semantic prototypes for fine-grained severity prediction. Furthermore, instance-level description preferences supervise fusion-stage attention, linking local semantic interaction with instance-level classification through consistent semantic guidance.

Our main contributions are summarized as follows:
\begin{itemize}
    \item We propose \textbf{Adaptive Text-Visual Alignment}, which enables evidence-adaptive interaction by contextualizing multiple damage descriptions with bi-temporal visual features under reliability-aware supervision.

    \item We develop \textbf{Dynamic Instance-Prototype Grounding}, which constructs instance-adaptive semantic prototypes for damage classification, with cross-stage supervision linking semantic fusion and instance prediction.

    \item Extensive experiments on xBD and BRIGHT demonstrate that LAMIA-IBDA achieves \textbf{state-of-the-art performance} in optical--optical and optical--SAR settings.
\end{itemize}

\section{Related Work}
\subsection{Building Damage Assessment}

Building damage assessment (BDA) has been extensively studied using bi-temporal RS imagery, with early approaches primarily focusing on pixel-level damage mapping. Following the introduction of the xBD benchmark~\cite{gupta2019xbd}, advances in RS change detection have enabled effective bi-temporal representation learning, including Siamese CNNs~\cite{daudt2018fully}, attention-based networks~\cite{chen2020stanet,fang2021snunet}, Transformers~\cite{chen2022bit,bandara2022changeformer}, and state-space models~\cite{chen2024changemamba}. Despite their effectiveness in dense prediction, pixel-level approaches do not explicitly preserve building identities~\cite{li2025overcoming,li2026progressive}. To address this limitation, recent studies have explored object-aware change modeling, including Mask View~\cite{ma2024maskview}, ChangeOS~\cite{zheng2021changeos}, and STS-STAM-CMR~\cite{yan2022instancechange}. Meanwhile, advances in instance segmentation~\cite{he2017maskrcnn,chen2019htc,cheng2022mask2former} provide effective foundations for building-level localization and delineation. The BRIGHT benchmark further extends instance-level BDA to heterogeneous optical--SAR imagery and diverse disaster scenarios~\cite{chen2025bright,chen2026brightchallenge}, highlighting the need for robust cross-modal damage representations. Nevertheless, damage severity estimation remains largely dependent on visual features and discrete category supervision. In contrast, our work incorporates explicit damage semantics into both bi-temporal feature fusion and instance-level classification.

\subsection{Vision--Language Alignment}

Vision--language learning has progressed from contextual language encoding and contrastive pretraining to multimodal interaction and prompt adaptation. BERT provides contextual language representations~\cite{devlin2019bert}, while CLIP and ALIGN establish visual--text alignment through contrastive pretraining~\cite{radford2021clip,jia2021align}. BLIP and BLIP-2 advance vision--language understanding and generation~\cite{li2022blip,li2023blip2}, while CoOp, CoCoOp, and MaPLe adapt pretrained representations through prompt learning~\cite{zhou2022coop,zhou2022cocoop,khattak2023maple}. For dense prediction, LSeg and DenseCLIP establish pixel--text correspondence, with DenseCLIP introducing image-conditioned prompting~\cite{li2022lseg,rao2022denseclip}; CAT-Seg aggregates image--text similarity costs for segmentation~\cite{cho2024catseg}. LAVT incorporates pixel--word interactions into visual encoding~\cite{yang2022lavt}, whereas Grounding DINO combines cross-modal feature enhancement with language-guided detection queries~\cite{liu2024groundingdino}. In RS, SkyScript, GeoRSCLIP, and RemoteCLIP develop domain-specific vision--language representations~\cite{wang2023skyscript,zhang2024georsclip,liu2024remoteclip}. For pixel-level change detection, ChangeCLIP introduces text semantics through a vision--language-driven decoder~\cite{dong2024changeclip}, MMChange fuses textual differences with visual features through attention~\cite{zhou2025mmchange}, and ViewDelta jointly models image, text, and segmentation query tokens for prompt-specified change prediction~\cite{varghese2025viewdelta}. BDAChat further combines segmentation with language-based damage reasoning~\cite{wang2026bdachat}. Building on these advances, LAMIA connects reliability-aware bi-temporal interaction with instance-level classification: ATVA contextualizes damage descriptions using local visual evidence, while DIPG aggregates text states into building-adaptive prototypes, linking language guidance across feature fusion and damage classification.

\begin{figure*}[t]
  \centering
  \includegraphics[width=\linewidth]{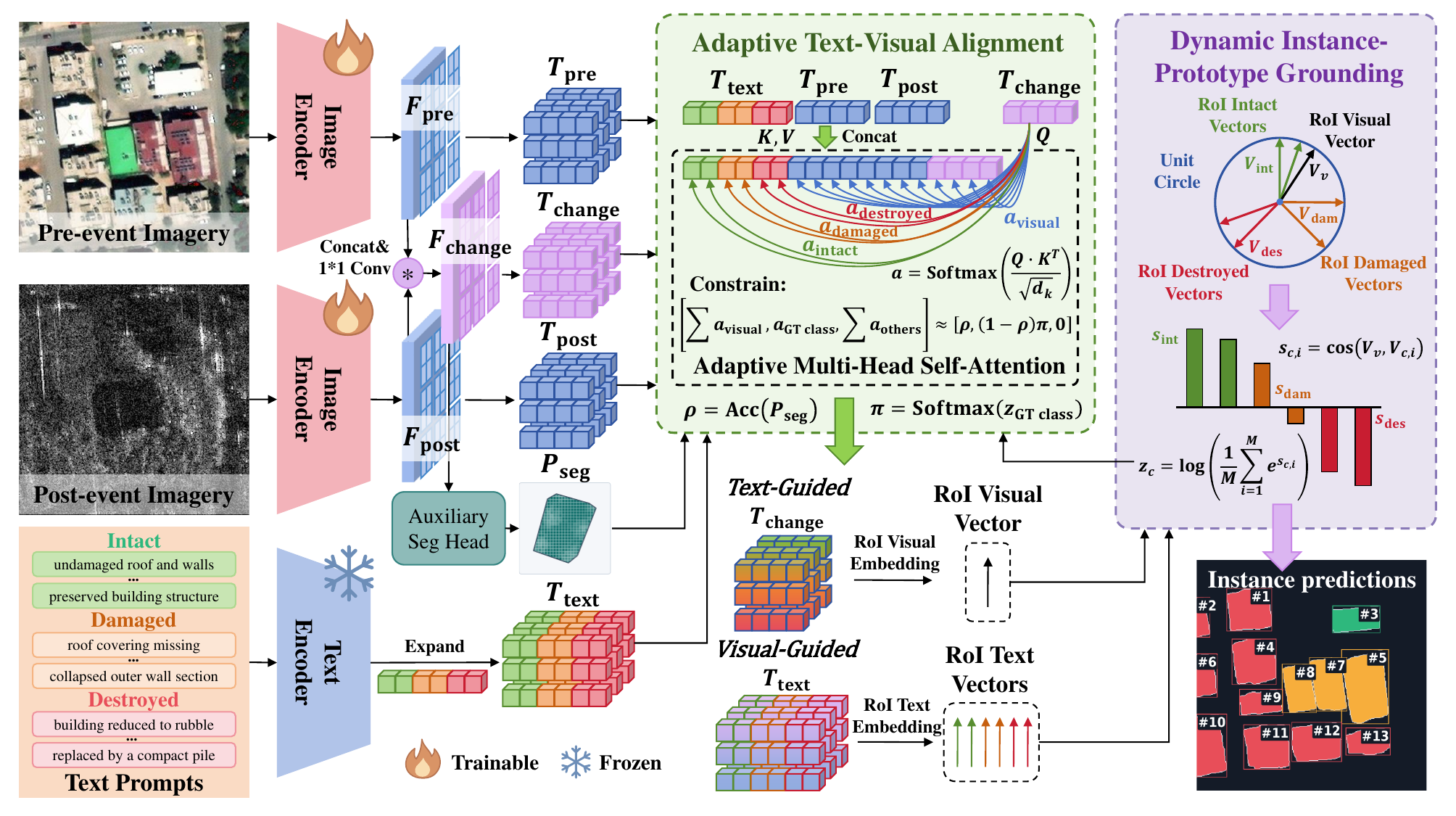}
\caption{Overview of LAMIA. ATVA adapts damage semantics to local bi-temporal evidence through reliability-aware interaction, while DIPG grounds contextualized semantics in individual buildings for damage classification. Instance-level description preferences further supervise fusion-stage attention.}
  \label{fig:model}
  \vspace{-1.0em}
\end{figure*}

\section{Methodology}
\label{sec:method}

\subsection{Overview}
\label{sec:overview}

Given aligned pre- and post-disaster images $I^{\mathrm{pre}}$ and $I^{\mathrm{post}}$, IBDA predicts building instances $\mathcal{Y}=\{(b_i,m_i,y_i)\}_{i=1}^{N}$, where $b_i$, $m_i$, and $y_i\in\{1,\ldots,C\}$ denote the bounding box, mask, and damage category. Each category $c$ is represented by multiple descriptions $\mathcal{D}_c=\{d_{c,k}\}_{k=1}^{K_c}$ to capture diverse damage patterns.

As shown in Fig.~\ref{fig:model}, LAMIA progressively grounds shared damage semantics in local bi-temporal evidence and individual building instances. \textbf{Adaptive Text-Visual Alignment} (ATVA) jointly contextualizes bi-temporal features and multiple damage descriptions under reliability-aware attention supervision. \textbf{Dynamic Instance-Prototype Grounding} (DIPG) further associates these contextualized semantics with candidate building regions to construct instance-adaptive prototypes for damage classification. Instance-level description preferences additionally supervise fusion-stage attention, linking local semantic adaptation with building-level discrimination. The framework employs an FPN~\cite{lin2017feature} and the three-stage cascade head of Hybrid Task Cascade (HTC)~\cite{chen2019htc} for instance detection and segmentation.

\subsection{Adaptive Text-Visual Alignment}
\label{sec:atva}

ATVA adapts multiple damage descriptions to local bi-temporal evidence through joint visual--language interaction, with attention supervision accounting for visual reliability and instance-specific semantic relevance.

\noindent\textbf{Visual-conditioned semantic interaction.}
At scale $s$, pre- and post-disaster features $F_s^{\mathrm{pre}}$ and $F_s^{\mathrm{post}}$ yield initial change features $G_s=\phi_s([F_s^{\mathrm{pre}};F_s^{\mathrm{post}}])$ through a $1\times1$ convolution. A frozen BGE encoder~\cite{BGE} and learnable projection produce independent description tokens $t_{c,k}=W_t\mathcal{E}_t(d_{c,k})$. Within each window $\Omega_{s,j}$, visual and text tokens undergo joint attention:
\begin{equation}
\begin{aligned}
Z_{s,j}^{0}&=[V_{s,j}^{\mathrm{pre}};V_{s,j}^{\mathrm{post}};V_{s,j}^{\Delta};T],\\
Z_{s,j}^{\ell+1}&=\mathcal{A}_{\ell}(Z_{s,j}^{\ell}),
\end{aligned}
\label{eq:joint_interaction}
\end{equation}
where $V_{s,j}^{\Delta}$ comes from $G_s$, $T=\{t_{c,k}\}_{c,k}$, and $\mathcal{A}_{\ell}$ denotes a joint attention block. The updated change features $\widehat G_{s,j}$ become semantically enriched, while text states $\widehat T_{s,j}$ are adapted to the local bi-temporal evidence.

\noindent\textbf{Reliability-adaptive semantic allocation.}
We supervise attention allocation at two complementary levels: the relative contribution of visual and textual evidence, and the relevance of individual damage descriptions. A visual-only auxiliary head $\mathcal H_v$ provides a training-time confidence proxy for ground-truth building $g$ at scale $s$:
\begin{equation}
\begin{aligned}
P_s^v&=\operatorname{softmax}(\mathcal H_v(G_s)),\\
\rho_{g,s}&=\operatorname{sg}\!\left[\frac{1}{|M_{g,s}|}\sum_{u\in M_{g,s}}P_s^v(u,y_g)\right],
\end{aligned}
\label{eq:visual_confidence}
\end{equation}
where $M_{g,s}$ and $y_g$ denote the building mask and true damage category, and $\operatorname{sg}$ stops gradients.

To capture description relevance, we aggregate the within-class similarity distributions of final-stage positive RoIs $\mathcal P(g)$ matched to building $g$:
\begin{equation}
\begin{aligned}
q_{i,k}&=\frac{\exp(s_{i,y_g,k}/\tau)}{\sum_{k'}\exp(s_{i,y_g,k'}/\tau)},\\
\pi_g&=\operatorname{sg}\!\left[\frac{1}{|\mathcal P(g)|}\sum_{i\in\mathcal P(g)}q_i\right],
\end{aligned}
\label{eq:description_preference}
\end{equation}
where $s_{i,c,k}$ is the instance--description similarity defined in Sec.~\ref{sec:dipg} and $\tau>0$ is a learnable temperature. We combine $\rho_{g,s}$ and $\pi_g$ into the target attention distribution
\begin{equation}
Q_{g,s}(a)=
\begin{cases}
\rho_{g,s}, & a=\mathcal V,\\
(1-\rho_{g,s})\pi_{g,k}, & a=t_{y_g,k},\\
0, & a=t_{c,k},\ c\neq y_g,
\end{cases}
\label{eq:semantic_allocation}
\end{equation}
where $\mathcal V$ aggregates all visual keys. Thus, visual confidence regulates the relative contribution of visual and textual evidence, while $\pi_g$ determines the semantic preference among descriptions of the ground-truth category.

Let $A_{g,s,\ell}$ denote change-query attention averaged over heads and building-mask locations, with visual keys grouped and text keys retained individually. We minimize
\begin{equation}
\mathcal L_{\mathrm{ATVA}}=
\sum_{s,\ell}\beta_s\gamma_\ell
\mathbb E_{g\in\mathcal G_s}
D_{\mathrm{KL}}(Q_{g,s}\|A_{g,s,\ell}),
\label{eq:atva_loss}
\end{equation}
where $\beta_s$ and $\gamma_\ell$ weight scales and layers. When $\mathcal P(g)$ is empty, only visual--text group-level supervision is applied. This objective guides attention without explicitly gating features and transfers instance-level description preferences back to fusion.

\subsection{Dynamic Instance-Prototype Grounding}
\label{sec:dipg}

ATVA produces locally contextualized text representations, which capture damage semantics from bi-temporal evidence but are not yet specific to individual buildings. DIPG further grounds these representations in candidate building regions to construct instance-adaptive semantic prototypes.

\begin{figure}[t]
  \centering
  \includegraphics[width=\linewidth]{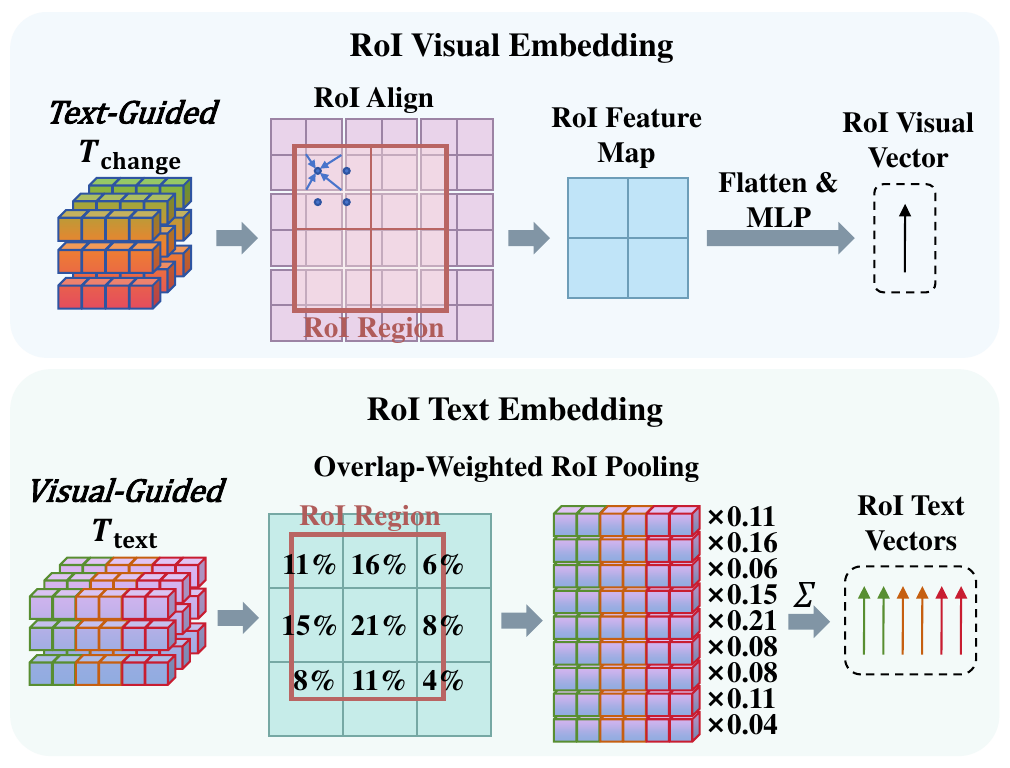}
  \caption{Instance-adaptive representations in DIPG. RoI visual features are matched with spatially aggregated text prototypes to construct building-specific semantic representations.}
  \label{fig:model_sub}
  \vspace{-1.0em}
\end{figure}

\noindent\textbf{Instance-conditioned semantic grounding.}
As illustrated in Fig.~\ref{fig:model_sub}, RoIAlign and an MLP extract visual embedding $x_i$ for candidate box $b_i$. Text states are pooled according to normalized spatial overlap:
\begin{equation}
\begin{aligned}
\alpha_{i,j}&=\frac{|b_i\cap\Omega_j|}{\sum_{j'}|b_i\cap\Omega_{j'}|},\\
p_{i,c,k}&=\sum_j\alpha_{i,j}\widehat T_{j,c,k},
\end{aligned}
\label{eq:contextual_prototype}
\end{equation}
where $\widehat T_{j,c,k}$ denotes the contextualized representation of description $k$ for category $c$ in window $\Omega_j$ from the selected feature scale. Overlap-based aggregation associates local damage semantics with each candidate building according to its spatial extent. Updating $\alpha_{i,j}$ with refined boxes allows the prototypes to track evolving instance regions.

\noindent\textbf{Evidence-adaptive description matching.}
We measure instance--prototype compatibility as $s_{i,c,k}=\cos(W_vx_i,p_{i,c,k})$, with learnable projection $W_v$. Rather than averaging prototypes, DIPG aggregates their similarities using
\begin{equation}
z_{i,c}=\log\!\left[\frac{1}{K_c}\sum_{k=1}^{K_c}\exp\!\left(\frac{s_{i,c,k}}{\tau}\right)\right].
\label{eq:prototype_score}
\end{equation}
This log-mean-exp score admits the equivalent variational interpretation
\begin{equation}
z_{i,c}=\max_{q\in\Delta^{K_c-1}}\left\{
\frac{1}{\tau}\mathbb E_{k\sim q}[s_{i,c,k}]
-D_{\mathrm{KL}}(q\|\mathcal U_c)\right\},
\label{eq:variational_matching}
\end{equation}
where $\mathcal U_c$ is uniform over category-$c$ descriptions. Its optimizer $q_{i,c,k}^*=\operatorname{softmax}_k(s_{i,c,k}/\tau)$ captures instance-specific description preferences while regularizing their concentration. These preferences are further used to supervise semantic attention in ATVA.

A visual foreground logit $o_i$ separates building presence from severity classification:
\begin{equation}
\begin{aligned}
P_i(\mathrm{bg})&=1-\sigma(o_i),\\
P_i(c)&=\sigma(o_i)\frac{\exp(z_{i,c})}{\sum_{c'=1}^{C}\exp(z_{i,c'})}.
\end{aligned}
\label{eq:instance_prediction}
\end{equation}


\subsection{Loss Function}
\label{sec:loss}

To further align instance-level visual representations with their contextualized damage semantics, we introduce an auxiliary alignment objective for positive RoIs. For positive RoI $i$ at cascade stage $r$, sample $\widehat k_i^{(r)}$ from the detached within-class distribution $q_{i,k}^{(r)}=\operatorname{softmax}_k(s_{i,y_i,k}^{(r)}/\tau)$. The alignment loss is
\begin{equation}
\mathcal L_{\mathrm{align}}=
\sum_{r=1}^{R}\omega_r
\mathbb E_{i\in\mathcal P_r}
\left[1-s_{i,y_i,\widehat k_i^{(r)}}^{(r)}\right],
\label{eq:alignment_loss}
\end{equation}
where $R=3$, $\mathcal P_r$ contains positive RoIs, and $\omega_r$ weights cascade stages. Sampling selects a description only for the auxiliary alignment loss; severity classification aggregates similarities over all descriptions.

The overall objective is
\begin{equation}
\begin{aligned}
\mathcal L_{\mathrm{total}}={}&\mathcal L_{\mathrm{inst}}
+\lambda_{\mathrm{sem}}\mathcal L_{\mathrm{sem}}^{\mathrm{aux}}\\
&+\lambda_{\mathrm{ATVA}}\mathcal L_{\mathrm{ATVA}}
+\lambda_{\mathrm{align}}\mathcal L_{\mathrm{align}},
\end{aligned}
\label{eq:total_loss}
\end{equation}
where $\mathcal L_{\mathrm{inst}}$ includes proposal, classification, box, and mask losses, and $\mathcal L_{\mathrm{sem}}^{\mathrm{aux}}$ supervises the visual-only head. At inference, the auxiliary visual head is discarded, and predictions are generated from paired images and predefined damage descriptions.

\section{Experiments}

\subsection{Experimental Setup}
\label{sec:exp_setup}

\begin{table*}[t]
  \centering
  \caption{Instance-level building damage assessment on BRIGHT and xBD test set. All AP values are on a 0--100 scale. Bold and underline indicate the best and second-best results, respectively.}
  \label{tab:comparison_matric}
  \setlength{\tabcolsep}{2pt}
  \renewcommand{\arraystretch}{1.08}
  \begin{tabular}{@{}lc*{18}{c}@{}}
    \toprule
    \multirow{3}{*}[-6pt]{Method} & \multicolumn{6}{c}{BRIGHT} & \; & \multicolumn{7}{c}{xBD} & &\multirow{3}{*}[-6pt]{\makecell{Params$\downarrow$\\(M)}} & \multirow{3}{*}[-6pt]{\makecell{FLOPs$\downarrow$\\(G)}} \\
    \cmidrule(lr){2-7}\cmidrule(lr){9-15}
    & \multicolumn{3}{c}{Mask AP$\uparrow$}
    & \multicolumn{3}{c}{Class-wise AP$\uparrow$} & & \multicolumn{3}{c}{Mask AP$\uparrow$} & \multicolumn{4}{c}{Class-wise AP$\uparrow$} \\
    \cmidrule(lr){2-4}\cmidrule(lr){5-7}\cmidrule(lr){9-11}\cmidrule(lr){12-15}
    & $\mathrm{AP}$ & $\mathrm{AP}_{50}$ & $\mathrm{AP}_{75}$ & Int. & Dam. & Des. & & $\mathrm{AP}$ & $\mathrm{AP}_{50}$ & $\mathrm{AP}_{75}$ & No-d. & Min. & Maj. & Des. \\
    \midrule
    \rowcolor[HTML]{C0C0C0}
    \multicolumn{18}{@{}l}{\textit{Pixel-level methods}} \\
    ChangeMamba~\cite{chen2024changemamba} & 11.1 & 22.8 & 10.0 & 14.9 & 0.8 & 17.6 & & 10.7 & 22.7 & 9.1 & 14.3 & 1.2 & 13.0 & 14.3 & &\underline{4.246} & \underline{17.171} \\
    Change3D~\cite{zhu2025change3d} & 11.3 & 23.0 & 10.1 & 16.2 & 1.8 & 15.8 & & 8.7 & 19.1 & 7.4 & 8.3 & 1.2 & 9.5 & 15.7 & & \textbf{0.125} & \textbf{15.477} \\
    CSDNet~\cite{dong2026csdnet} & 12.4 & 24.1 & 11.8 & 17.0 & 1.1 & 19.0 & & 11.8 & 24.3 & 10.7 & 17.4 & 2.4 & 11.7 & 15.8 & &9.499 & 220.386 \\
    \midrule
    \rowcolor[HTML]{C0C0C0}
    \multicolumn{18}{@{}l}{\textit{Instance-level methods}} \\
    Mask R-CNN~\cite{he2017maskrcnn} & 28.7 & 49.3 & 30.3 & 39.1 & 13.2 & 33.8 & & 28.6 & 51.1 & 30.2 & 35.0 & 17.3 & 30.4 & 31.9 & &31.628 & 168.055 \\
    Cas. Mask R-CNN~\cite{cai2018cascade} & 28.8 & 47.9 & 30.9 & 40.6 & 11.4 & 34.4 & &29.4 & 50.9 & 31.7 & 35.4 & 17.4 & 31.3 & 33.5 & &64.673 & 358.583 \\
    HTC~\cite{chen2019htc} & \underline{30.8} & 51.4 & \underline{32.8} & 41.1 & 14.7 & \underline{36.7} & & \underline{30.8} & 53.9 & \underline{32.7} & 36.5 & \underline{19.1} & \underline{32.8} & \underline{34.7} & &67.561 & 386.280 \\
    SOLOv2~\cite{wang2020solov2} & 26.2 & 47.8 & 26.7 & 36.1 & 11.3 & 31.2 & &26.0 & 50.3 & 25.1 & 31.1 & 14.7 & 29.2 & 29.0 & &33.883 & 200.909 \\
    SCNet~\cite{vu2021scnet} & 30.7 & 51.9 & 32.5 & 41.1 & \underline{15.4} & 35.5 & &30.1 & 53.0 & 31.7 & 35.9 & 18.0 & \underline{32.8} & 33.8 & &82.122 & 386.904 \\
    Mask2Former~\cite{cheng2022mask2former} & 28.3 & 49.5 & 29.1 & 39.6 & 12.6 & 32.6 & &29.4 & 53.4 & 30.3 & 35.2 & 18.0 & 30.9 & 33.6 & &31.849 & 55.895 \\
    FastInst~\cite{he2023fastinst} & 25.4 & 47.4 & 24.7 & 35.8 & 10.5 & 29.9 & &25.0 & 48.6 & 24.4 & 31.3 & 12.7 & 26.7 & 29.4 & &21.740 & 39.999 \\
    ContourFormer~\cite{yao2025contourformer} & 29.3 & 50.2 & 30.6 & \underline{41.6} & 12.6 & 33.7 & &29.8 & 54.7 & 30.7 & 37.1 & 18.5 & 30.6 & 33.0 & &29.315 & 102.116 \\
    D-FINE-seg~\cite{saakyan2026dfineseg} & 30.1 & \underline{52.1} & 31.5 & 41.5 & 14.6 & 34.2 & &30.1 & \underline{55.6} & 30.5 & \underline{37.2} & 18.4 & 31.2 & 33.6 & &30.174 & 73.371 \\
    \midrule
    \rowcolor[HTML]{C0C0C0}
    \multicolumn{18}{@{}l}{\textit{Text-guided instance-level methods}} \\
    LAMIA (Ours)
    & \textbf{33.5} & \textbf{55.4} & \textbf{36.1} & \textbf{43.1} & \textbf{17.4} & \textbf{39.9} & &\textbf{33.4} & \textbf{57.1} & \textbf{36.2} & \textbf{39.0} & \textbf{21.7} & \textbf{35.7} & \textbf{37.3} & & 58.985 & 369.881 \\
    \bottomrule
  \end{tabular}
\end{table*}

LAMIA adopts a bi-temporal PVT-v2 backbone~\cite{wang2022pvt}, FPN~\cite{lin2017feature}, and a three-stage cascade head that interleaves bounding-box refinement and instance mask prediction~\cite{chen2019htc}. LAMIA employs multiple descriptions per damage class, encoded by a frozen BGE encoder~\cite{BGE}. All models are trained for 100 epochs using AdamW~\cite{loshchilov2017decoupled} with an initial learning rate of $10^{-4}$. LAMIA is trained with a batch size of 8, random $512\times512$ crops, and random flipping. During inference, each $1024\times1024$ image is divided into four quadrants, with predictions mapped back to the original image coordinates. We report COCO-style mask AP~\cite{lin2014microsoft} averaged over IoU thresholds from 0.50 to 0.95 at intervals of 0.05, along with $\mathrm{AP}_{50}$, $\mathrm{AP}_{75}$ and class-wise mask AP. All metrics are reported on a 0--100 scale.

\subsection{Datasets}
\label{sec:exp_datasets}

\noindent\textbf{BRIGHT.}
BRIGHT~\cite{chen2025bright} comprises paired pre-disaster optical and post-disaster SAR imagery covering diverse disaster events, including earthquakes, wildfires, and explosions, etc. It contains 2,332 training pairs with 186,373 building instances and 697 test pairs with 58,579 instances.

\noindent\textbf{xBD.}
xBD~\cite{gupta2019xbd} pairs pre- and post-disaster optical imagery across diverse disaster events, including floods, tsunamis, and volcanic eruptions, etc. It contains 2,799 training pairs with 159,794 building instances and 933 test pairs with 53,137 instances.



\subsection{Comparison Methods}
\label{sec:exp_methods}

We compare LAMIA with three pixel-level methods~\cite{chen2024changemamba,zhu2025change3d,dong2026csdnet}
and nine instance-level methods~\cite{he2017maskrcnn,cai2018cascade,chen2019htc,wang2020solov2,vu2021scnet,cheng2022mask2former,he2023fastinst,yao2025contourformer,saakyan2026dfineseg},
covering two-stage, cascaded, query-based, and lightweight architectures.
To support instance-level damage assessment, we post-process pixel-level
predictions into individual building instances with damage labels.
Single-image instance models are extended to Siamese architectures that
extract and fuse pre- and post-disaster features for instance prediction.



\subsection{Quantitative Results}
\label{sec:quantitative}

As shown in Tab.~\ref{tab:comparison_matric}, LAMIA achieves mask AP scores of 33.5 on BRIGHT and 33.4 on xBD, surpassing the strongest baselines by 2.7 and 2.6 points, with larger gains of 3.3 and 3.5 points at $\mathrm{AP}_{75}$. It achieves the highest class-wise AP across all damage categories, improving damaged-building AP on BRIGHT by 2.0 points and minor-damage AP on xBD by 2.6 points. Compared with its underlying instance segmentation framework, HTC~\cite{chen2019htc}, LAMIA reduces parameters from 67.6M to 59.0M and FLOPs from 386.3G to 369.9G while improving accuracy. 

\subsection{Qualitative Results}
\label{sec:qualitative}

\begin{figure*}[t]
  \centering
  \includegraphics[width=\linewidth]{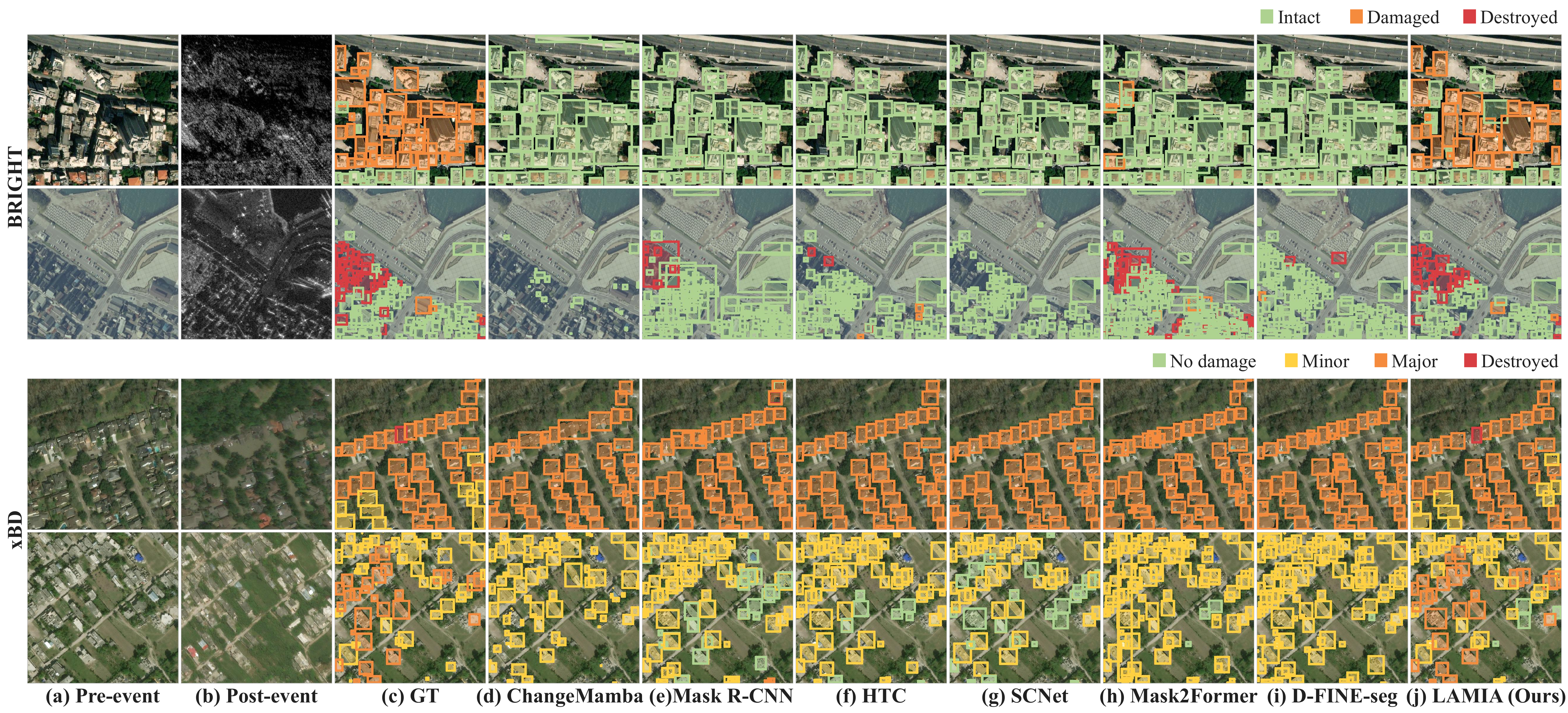}
  \caption{Qualitative comparison of building damage assessment on BRIGHT and xBD test set. Colors indicate building damage grades.}
  \label{fig:comparison}
  \vspace{-0.5em}
\end{figure*}

Fig.~\ref{fig:comparison} compares qualitative results on BRIGHT and xBD. On BRIGHT, LAMIA better identifies damaged and destroyed buildings misclassified as intact by visual baselines. On xBD, it more accurately distinguishes minor and major damage, demonstrating improved discrimination between similar damage grades. We further assess generalization to unseen scenes through large-area predictions for the 2011 Joplin tornado and 2025 Eaton Fire in Altadena, California, with additional results in the \textbf{Supplementary Material}.

\section{Ablation Study}
\label{sec:ablations}

We conduct ablation studies on BRIGHT and xBD to evaluate the proposed components and language-guided strategies, including dynamic semantic adaptation and instance-level grounding. All ablation experiments follow the training and evaluation protocol described in Sec.~\ref{sec:exp_setup}, with each variant retrained under its respective configuration. Further experimental results and analyses are presented in the \textbf{Supplementary Material}.


\subsection{Component Contributions}
\label{sec:ablation_components}

\begin{figure*}[t]
  \centering
  \includegraphics[width=\linewidth]{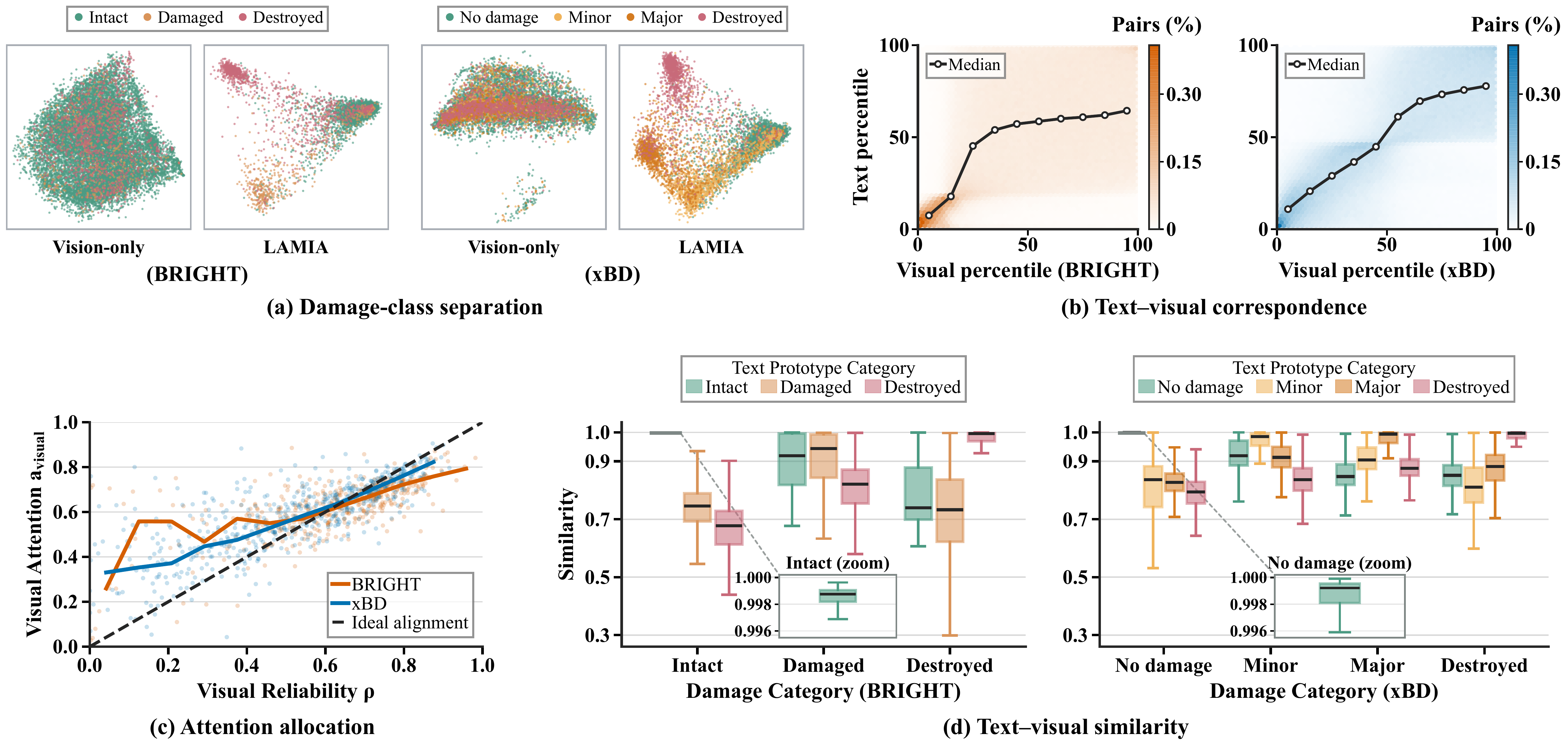}
  \caption{Visual representations, text--visual correspondence, attention allocation,
  and semantic matching on BRIGHT and xBD. Pairwise similarities use global
  percentiles, and curves show binned medians. Box plots report LSE-aggregated
  cosine scores; boxes show interquartile ranges with
  $1.5\times\mathrm{IQR}$ whiskers.}
  \label{fig:representation_analysis}
  \vspace{-1em}
\end{figure*}

Tab.~\ref{tab:core_ablation} shows that combining ATVA and DIPG achieves the best performance on both datasets, improving mask AP over the vision-only baseline by 2.0 points on BRIGHT and 1.7 points on xBD. Using ATVA or DIPG alone yields smaller gains of 1.5/1.3 and 1.1/1.2 points, respectively. The consistent advantage of the full model supports our progressive design: ATVA adapts damage semantics to local bi-temporal evidence, and DIPG grounds these contextualized semantics in individual buildings for severity classification. Together, they connect semantic adaptation during feature fusion with instance-level discrimination.
Fig.~\ref{fig:representation_analysis}(a) further visualizes GT RoI embedding distributions using PCA. Compared with the vision-only baseline, LAMIA produces more distinct groupings across damage grades, indicating improved visual discrimination through language guidance.


\begin{table}[t]
  \centering
  \caption{Ablation study of the proposed components on BRIGHT and xBD.}
  \label{tab:core_ablation}
  \setlength{\tabcolsep}{2.5pt}
  \renewcommand{\arraystretch}{1.12}
  \begin{tabular}{@{}ccc*{6}{c}@{}}
    \toprule
    \multirow{2}{*}[-3pt]{Vision}
    & \multirow{2}{*}[-3pt]{ATVA}
    & \multirow{2}{*}[-3pt]{DIPG}
    & \multicolumn{3}{c}{BRIGHT}
    & \multicolumn{3}{c}{xBD} \\
    \cmidrule(lr){4-6}\cmidrule(lr){7-9}
    & & & AP & AP$_{50}$ & AP$_{75}$
    & AP & AP$_{50}$ & AP$_{75}$ \\
    \midrule
    $\checkmark$ & & 
    & 31.5 & 52.7 & 33.9
    & 31.7 & 55.0 & 34.4 \\
    $\checkmark$ & $\checkmark$ &
    & \underline{33.0} & \underline{53.8} & \underline{35.9}
    & \underline{33.0} & \underline{56.7} & \underline{35.6} \\
    $\checkmark$ & & $\checkmark$
    & 32.6 & 52.7 & 35.3
    & 32.9 & 56.5 & \underline{35.6} \\
    \midrule
    $\checkmark$ & $\checkmark$ & $\checkmark$
    & \textbf{33.5} & \textbf{55.4} & \textbf{36.1}
    & \textbf{33.4} & \textbf{57.1} & \textbf{36.2} \\
    \bottomrule
  \end{tabular}
\end{table}

\subsection{Comparison of Language-guided Strategies}
\label{sec:ablation_text_comparison}

Tab.~\ref{tab:text_comparison} compares LAMIA with six alternative language-integration strategies under the same bi-temporal instance framework. We adapt their fusion or prompt-learning mechanisms while preserving their characteristic approaches to language-guided feature learning or region classification, denoted as \emph{method-style} adaptations. Existing strategies yield inconsistent gains, with some degrading performance. In contrast, LAMIA achieves the highest mask AP on BRIGHT and xBD, surpassing the strongest alternatives by 1.6 and 1.4 points, respectively. These results highlight the benefit of jointly adapting damage semantics to local bi-temporal evidence and individual buildings for reliable damage classification.

\begin{table}[t]
  \centering
  \caption{Comparison of different language-guided strategies on BRIGHT and xBD.}
  \label{tab:text_comparison}
  \setlength{\tabcolsep}{2pt}
  \renewcommand{\arraystretch}{1.12}
  \begin{tabular}{@{}lcccccc@{}}
    \toprule
    \multirow{2}{*}[-3pt]{Method}
    & \multicolumn{3}{c}{BRIGHT}
    & \multicolumn{3}{c}{xBD} \\
    \cmidrule(lr){2-4}\cmidrule(lr){5-7}
    & AP & AP$_{50}$ & AP$_{75}$
    & AP & AP$_{50}$ & AP$_{75}$ \\
    \midrule
    Vision-only
    & 31.5 & \underline{52.7} & 33.9
    & 31.7 & 55.0 & \underline{34.4} \\
    DenseCLIP-style~\cite{rao2022denseclip}
    & 31.1 & 50.7 & 31.9
    & 31.7 & \textbf{58.2} & 33.2 \\
    GLIP-style~\cite{li2022grounded}
    & \underline{31.9} & 49.9 & 34.5
    & 30.6 & 53.9 & 33.9 \\
    LAVT-style~\cite{yang2022lavt}
    & 30.6 & 49.4 & 31.9
    & \underline{32.0} & \underline{57.7} & 34.2 \\
    ChangeCLIP-style~\cite{dong2024changeclip}
    & 30.3 & 48.6 & 31.7
    & 30.6 & 55.3 & 32.8 \\
    MMChange-style~\cite{zhou2025mmchange}
    & 30.4 & 49.6 & 31.1
    & 31.4 & 56.8 & 33.7 \\
    CoCoOP-style~\cite{zhou2022cocoop}
    & 31.6 & 49.8 & \underline{34.6}
    & 31.0 & 55.5 & 33.6 \\
    \midrule
    LAMIA (Ours)
    & \textbf{33.5} & \textbf{55.4} & \textbf{36.1}
    & \textbf{33.4} & 57.1 & \textbf{36.2} \\
    \bottomrule
  \end{tabular}
\end{table}

\subsection{Dynamic Adaptation of Damage Semantics}
\label{sec:ablation_contextualization}

To evaluate dynamic semantic adaptation, we ablate both text contextualization and adaptive attention supervision. As shown in Tab.~\ref{tab:inference_ablations}, replacing evidence-adaptive text updates with frozen text representations reduces AP from 33.5 to 30.6 on BRIGHT and from 33.4 to 32.0 on xBD, highlighting the importance of adapting damage semantics in visual evidence. Fixing visual reliability at $\rho_{g,s}=0.5$ or replacing instance-level description preferences with a uniform distribution $\pi_{g,k}=1/K_{y_g}$ reduces AP in both cases. These results further support reliability-aware allocation between visual and textual evidence and instance-specific weighting of damage descriptions.
Fig.~\ref{fig:representation_analysis}(b) shows a positive correlation between visual and contextual-text similarities across GT building pairs, suggesting that contextualized text captures instance-specific appearance. Fig.~\ref{fig:representation_analysis}(c) shows that visual attention increases with reliability, with language receiving greater weight when visual evidence is less reliable.

\begin{table}[t]
  \centering
\caption{Impact of dynamic text representations and semantic alignment on instance-level building damage assessment.}
  \label{tab:inference_ablations}
  \setlength{\tabcolsep}{2.5pt}
  \renewcommand{\arraystretch}{1.12}
  \begin{tabular}{@{}lcccccc@{}}
    \toprule
    \multirow{2}{*}[-3pt]{Variant}
    & \multicolumn{3}{c}{BRIGHT}
    & \multicolumn{3}{c}{xBD} \\
    \cmidrule(lr){2-4}\cmidrule(lr){5-7}
    & AP & AP$_{50}$ & AP$_{75}$
    & AP & AP$_{50}$ & AP$_{75}$ \\
    \midrule
    Frozen Text
    & 30.6 & 49.8 & 33.0
    & 32.0 & 53.7 & 34.7 \\
    Fixed Reliability
    & 30.8 & 50.6 & 33.8
    & 31.6 & 51.8 & 34.0 \\
    Uniform Preferences
    & \underline{31.5} & 51.8 & \underline{34.0}
    & 32.0 & 54.3 & 35.1\\
    Static Prototypes
    & 30.4 & 49.7 & 33.0
    & 31.9 & 53.6 & 34.6 \\
    Random Embeddings
    & 31.4 & 52.1 & 33.9
    & \underline{32.2} & 55.1 & \underline{35.3} \\
    Repeat Prompts
    & 31.2 & \underline{52.6} & 33.7
    & 31.9 & \underline{55.4} & 34.6 \\
    \midrule
    Full LAMIA
    & \textbf{33.5} & \textbf{55.4} & \textbf{36.1}
    & \textbf{33.4} & \textbf{57.1} & \textbf{36.2} \\
    \bottomrule
  \end{tabular}
\end{table}

\subsection{Instance-adaptive Semantic Grounding}
\label{sec:ablation_grounding}

\begin{figure}[t]
  \centering
  \includegraphics[width=\linewidth]{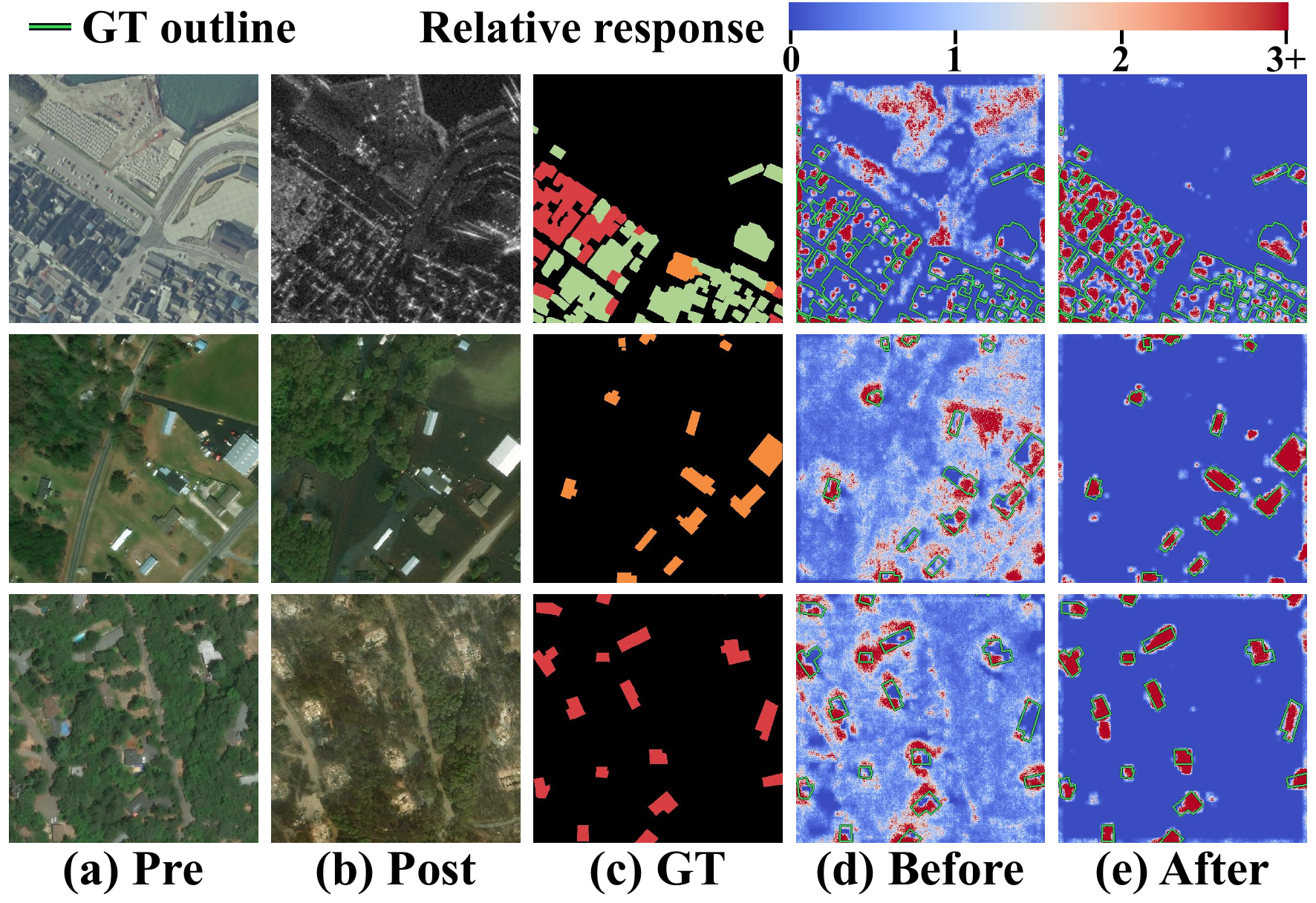}
\caption{Spatial responses before and after text contextualization. }
  \label{fig:text_response}
  \vspace{-1.5em}
\end{figure}

We evaluate the importance of instance-level semantic grounding. Replacing contextualized prototypes with static text embeddings reduces AP by 3.1 points on BRIGHT and 1.5 points on xBD (Tab.~\ref{tab:inference_ablations}), highlighting the benefit of instance-adaptive semantic references.
Replacing damage descriptions with random text embeddings or an equal number of repeated class-name prompts causes smaller AP drops, indicating that both semantic content and within-class diversity are beneficial, while instance-adaptive semantic alignment partially offsets their absence.
Fig.~\ref{fig:representation_analysis}(d) further evaluates instance--prototype alignment, after log-mean-exp aggregation of cosine similarities, matching-category prototypes yield the highest median scores across GT RoI groups on both datasets. Fig.~\ref{fig:text_response} shows that text contextualization concentrates responses within buildings and suppresses background activations, indicating improved semantic correspondence with building-level visual evidence.

\section{Conclusion}

We presented LAMIA-IBDA, a language-guided framework for instance-level building damage assessment. ATVA and DIPG adapts shared damage semantics to local bi-temporal evidence and individual buildings, enabling reliability-aware feature fusion and instance-adaptive classification. Consistent gains on BRIGHT and xBD across optical--SAR and optical--optical settings demonstrate the effectiveness of this progressive semantic grounding for fine-grained damage assessment.


{
    \small
    \section*{LLM Usage Statement}

We used OpenAI's GPT-6.1-Sol model to assist with English translation, grammar checking, and language polishing, as well as code refinement to improve robustness and computational efficiency. LLMs were not used for any other purposes.
    \bibliographystyle{ieeenat_fullname}
    \bibliography{main}
}
\clearpage
\setcounter{page}{1}
\maketitlesupplementary

\appendix

\section{Implementation Details}
\label{sec:supp-implementation}

LAMIA follows the same implementation on BRIGHT and xBD, with dataset-specific input streams, damage categories, and description banks.

\subsection{Model Architecture}
\label{sec:supp-backbone}

\noindent\textbf{Backbone and FPN.}
At pyramid level $s$, bi-temporal features are extracted and fused as
\begin{equation}
\begin{aligned}
F_s^{\mathrm{pre}} &= B_{\mathrm{pre}}(I^{\mathrm{pre}})_s,\\
F_s^{\mathrm{post}} &= B_{\mathrm{post}}(I^{\mathrm{post}})_s,\\
G_s &= \phi_s\!\left([F_s^{\mathrm{pre}};F_s^{\mathrm{post}}]\right),
\end{aligned}
\label{eq:supp-bi-temporal-fusion}
\end{equation}
where $\phi_s$ denotes a bias-free $1\times1$ convolution. Both streams use PVT-v2-B3 encoders~\cite{wang2022pvt}, producing four feature levels with channels $(64,128,320,512)$ and strides $(4,8,16,32)$. BRIGHT employs separate encoders for pre-event optical and post-event SAR imagery, whereas xBD shares encoder weights between its two optical streams. Therefore, we exclude the backbone from all reported parameters and FLOPs to ensure consistent complexity accounting across datasets.

PVT-v2-B3 has stage depths $(3,4,18,3)$, attention heads $(1,2,5,8)$, spatial-reduction ratios $(8,4,2,1)$, MLP ratios $(8,8,4,4)$, and a maximum drop-path rate of $0.1$. The encoders are initialized from pretrained weights. For BRIGHT, the SAR input is replicated to three channels, with the first convolution initialized by averaging the optical input-channel weights:
\begin{equation}
W_{\mathrm{SAR}}[:,q,:,:]
=\frac{1}{3}\sum_{q'=1}^{3}W_{\mathrm{opt}}[:,q',:,:],
\quad q\in\{1,2,3\}.
\label{eq:supp-sar-init}
\end{equation}

The fused features are projected to $d=128$ for ATVA, followed by a five-level, 256-channel FPN and a three-stage HTC head. Each box head uses $7\times7$ RoI features~\cite{he2017maskrcnn} and two 1024-dimensional fully connected layers with class-agnostic regression. Each mask head contains four $3\times3$ convolutions with 256 channels. The HTC stage weights are $(1,0.5,0.25)$.

\noindent\textbf{Frozen description encoding.}
Each damage description $d_{c,k}$ is encoded using a frozen BGE-small encoder~\cite{BGE}. Its first-token embedding is normalized and projected into the shared ATVA/DIPG space:
\begin{equation}
\begin{aligned}
e_{c,k} &=
\frac{E_{\mathrm{BGE}}(d_{c,k})}
{\|E_{\mathrm{BGE}}(d_{c,k})\|_2},\\
t_{c,k} &= \operatorname{LN}(W_t e_{c,k}).
\end{aligned}
\label{eq:supp-text-projection}
\end{equation}
The frozen BGE embeddings are computed once and reused across images and ATVA windows; the trainable projection is updated during training.

\noindent\textbf{Local ATVA.}
At each backbone level, projected features are partitioned into non-overlapping $8\times8$ windows~\cite{liu2021swin}. For window $\Omega_{s,j}$, the input sequence is
\begin{equation}
\begin{aligned}
Z_{s,j}^{0}
&=\left[V_{s,j}^{\mathrm{pre}};
V_{s,j}^{\mathrm{post}};
V_{s,j}^{\Delta};T\right],\\
T&=\{t_{c,k}\}_{c,k}.
\end{aligned}
\label{eq:supp-atva-tokens}
\end{equation}
The three visual token groups receive normalized global 2-D sine/cosine positional encodings~\cite{carion2020end}, while all groups receive learned type embeddings. Each pre-LN attention--FFN block is defined as
\begin{equation}
\begin{aligned}
U_{s,j}^{\ell}
&=Z_{s,j}^{\ell-1}
+\operatorname{MHA}\!\left(
\operatorname{LN}(Z_{s,j}^{\ell-1});M_{s,j}
\right),\\
Z_{s,j}^{\ell}
&=U_{s,j}^{\ell}
+\operatorname{FFN}\!\left(
\operatorname{LN}(U_{s,j}^{\ell})
\right).
\end{aligned}
\label{eq:supp-atva-block}
\end{equation}
ATVA uses three blocks with four attention heads each. The mask $M_{s,j}$ excludes padded keys in boundary windows.

Attention weights for auxiliary supervision are extracted from change-token queries before attention dropout:
\begin{equation}
A_{qk}=\operatorname{softmax}_{k}\!\left(
\frac{Q_qK_k^{\mathsf T}}{\sqrt{d_h}}+M_{qk}
\right).
\label{eq:supp-attention}
\end{equation}
Visual keys are grouped for supervision, while individual descriptions retain separate attention weights. The updated change tokens are projected back to the feature channels:
\begin{equation}
\widehat G_{s,j}=\phi_s^{\mathrm{out}}
\!\left(\operatorname{LN}(V_{s,j}^{\Delta,L})\right).
\label{eq:supp-atva-output}
\end{equation}
Here, $V_{s,j}^{\Delta}$ is obtained from the fused change feature
$G_s$ within $\Omega_{s,j}$. The contextualized text states
$\widehat T_{s,j}$ are retained for subsequent RoI stages.

\noindent\textbf{Reliability-aware attention supervision.}
A training-only visual auxiliary head $H_v$ predicts $C+1$ semantic
channels from each $G_s$:
\begin{equation}
\begin{aligned}
P_s^v
&=\operatorname{softmax}\!\left(H_v(G_s)\right).
\end{aligned}
\label{eq:supp-semantic-probe}
\end{equation}
For a ground-truth building $g$ with mask $M_{g,s}$ at scale $s$ and
damage class $y_g$, the visual confidence is defined as
\begin{equation}
\rho_{g,s}=\operatorname{sg}\!\left[
\frac{1}{|M_{g,s}|}
\sum_{u\in M_{g,s}}
P^v_s(u,y_g)\right].
\label{eq:supp-reliability}
\end{equation}
The final-stage positive RoIs provide a detached description preference,
normalized within the ground-truth class:
\begin{equation}
\begin{aligned}
\pi_{g,k}&=\operatorname{sg}\!\left[
\frac{1}{|\mathcal P(g)|}
\sum_{i\in\mathcal P(g)}q_{i,k}\right],
\\
q_{i,k}&=\operatorname{softmax}_k\!\left(s_{i,y_g,k}/\tau\right).
\end{aligned}
\label{eq:supp-description-preference}
\end{equation}
The target attention distribution assigns mass $\rho_{g,s}$ to visual
keys and mass $1-\rho_{g,s}$ to the corresponding damage descriptions:
\begin{equation}
Q_{g,s}(a)=
\begin{cases}
\rho_{g,s}, & a\in\mathcal V,\\
(1-\rho_{g,s})\,\pi_{g,k}, & a=t_{y_g,k},\\
0, & a=t_{c,k},\ c\ne y_g.
\end{cases}
\label{eq:supp-attn-target}
\end{equation}
Here, $\mathcal V$ denotes the aggregated visual-key group. Thus,
lower visual confidence increases the relative allocation to language
descriptions. Let $A_{g,s,\ell}$ denote the head-averaged, mask-pooled
change-query attention at block $\ell$. The attention supervision loss is
\begin{equation}
\begin{aligned}
\mathcal L_{\mathrm{ATVA}}
&=\sum_{s=1}^{3}\sum_{\ell=1}^{3}
\beta_s\gamma_\ell\,\overline D_{s,\ell},\\
\overline D_{s,\ell}
&=\frac{1}{|\mathcal G_s|}
\sum_{g\in\mathcal G_s}
D_{\mathrm{KL}}(Q_{g,s}\|A_{g,s,\ell}),\\[-1pt]
\beta&=(0.6,0.3,0.1),\quad
\gamma=(0.1,0.3,0.6).
\end{aligned}
\label{eq:supp-atva-loss}
\end{equation}
The weights emphasize finer spatial scales and deeper attention blocks.
The visual auxiliary head is removed at inference.

\noindent\textbf{DIPG and cascade classification.}
At HTC stage $r$, RoIAlign and an MLP extract a visual embedding $x_i$
for each proposal $b_i$. At the fixed feature-pyramid level used for RoI classification, we aggregate the contextualized text state from each ATVA window according to the candidate box’s detached area-overlap weights with those windows:
\begin{equation}
\begin{aligned}
\alpha_{ij}
&=\frac{|b_i\cap\Omega_j|}
{\sum_{j'}|b_i\cap\Omega_{j'}|},\\
p_{i,c,k}
&=\sum_j\alpha_{ij}\widehat T_{j,c,k}.
\end{aligned}
\label{eq:supp-roi-pooling}
\end{equation}
At each cascade stage, we compute overlap weights using the current proposal boxes. A valid proposal aggregates only the contextualized text states from the ATVA windows that intersect it, weighted by normalized intersection areas. If a proposal has no valid image area after clipping, we use a fallback average weighted by the valid image area of each window.

After linear projection and $\ell_2$ normalization, visual and text
features are compared using
\begin{equation}
s_{i,c,k}=\cos\!\left(W_vx_i,p_{i,c,k}\right).
\label{eq:supp-instance-similarity}
\end{equation}
Class scores are aggregated by log-mean-exp:
\begin{equation}
z_{i,c}
=\log\!\left[
\frac{1}{K_c}\sum_{k=1}^{K_c}
\exp\!\left(\frac{s_{i,c,k}}{\tau}\right)
\right],
\qquad \tau>0,
\label{eq:supp-logmeanexp}
\end{equation}
where $\tau$ is a learnable temperature shared with the
description-preference distribution.
An independent logit $o_i$ estimates building presence. The resulting
foreground and background probabilities are
\begin{equation}
\begin{aligned}
P_i(\mathrm{bg})&=1-\sigma(o_i),\\
P_i(c)&=\sigma(o_i)
\frac{\exp(z_{i,c})}{\sum_{c'=1}^{C}\exp(z_{i,c'})}.
\end{aligned}
\label{eq:supp-foreground-score}
\end{equation}

For each positive RoI, the alignment loss samples a description from the detached same-class distribution
\begin{equation}
q_{i,k}^{(r)}
=\operatorname{softmax}_k
\!\left(s_{i,y_i,k}^{(r)}/\tau\right).
\label{eq:supp-align-sampling}
\end{equation}
The resulting loss is
\begin{equation}
\begin{aligned}
\mathcal L_{\mathrm{align}}
&=\sum_{r=1}^{3}\omega_r\,\overline\ell_r,\\
\overline\ell_r
&=\frac{1}{|\mathcal P_r|}
\sum_{i\in\mathcal P_r}
\left(1-s_{i,y_i,\hat k_i^{(r)}}^{(r)}\right),\\[-1pt]
\omega&=(1,0.5,0.25).
\end{aligned}
\label{eq:supp-align-loss}
\end{equation}
where the sampled description index $\hat k_i^{(r)}$ is detached from back-propagation.

\subsection{Training and Inference}
\label{sec:supp-optimization}

The training objective combines instance-level HTC losses with semantic,
attention, and alignment supervision:
\begin{equation}
\mathcal L_{\mathrm{total}}=\mathcal L_{\mathrm{inst}}
+0.2\,\mathcal L_{\mathrm{sem}}^{\mathrm{aux}}
+0.1\,\mathcal L_{\mathrm{ATVA}}
+\mathcal L_{\mathrm{align}}.
\label{eq:supp-total-loss}
\end{equation}
Here, $\mathcal L_{\mathrm{inst}}$ contains the proposal,
classification, bounding-box, and mask losses, while
$\mathcal L_{\mathrm{sem}}^{\mathrm{aux}}$ supervises the training-only
visual auxiliary head.

\noindent\textbf{Optimization.}
We train for 100 epochs using AdamW with an initial learning rate of $10^{-4}$, $\beta=(0.9,0.999)$, $\epsilon=10^{-8}$, and weight decay $0.05$ (zero for normalization layers). Gradients are clipped at a norm of $10$. The learning rate follows a 2-epoch linear warm-up starting at $0.1$ of the initial rate, followed by cosine annealing from epochs 50 to 100 with a minimum rate of $0.5$ of the initial rate. Backbone learning-rate multipliers are $0.1$ for optical encoders and $0.3$ for the BRIGHT SAR encoder. We use a random seed of $42$, a per-process batch size of $2$ with $4$ NVIDIA RTX 3090 GPUs.

\noindent\textbf{Data preparation and inference.}
Training uses random $512\times512$ crops from $1024\times1024$ scenes, discarding instances smaller than one pixel after cropping. Horizontal, vertical, and diagonal flips are applied with probability $0.5$.
At evaluation time, each test image is divided into four fixed
$512\times512$ tiles. We process all four tiles using
$2{,}500$ pre-NMS proposals, $1{,}500$ post-NMS proposals, and at most
$500$ instances per patch. Predictions are mapped back to the original
$1024\times1024$ coordinates for COCO-style evaluation. At inference,
the frozen text encoder, visual auxiliary head, and auxiliary losses are
inactive; only the pre-encoded description bank, ATVA, and DIPG are used.

\section{Experimental Setup}

\subsection{Datasets and Evaluation Protocol}
\label{sec:supp_dataset_details}

We evaluate instance-level building damage assessment on BRIGHT (pre-event optical and post-event SAR) and xBD (bi-temporal optical). All $1024\times1024$ image pairs and annotations are organized in COCO instance segmentation format. Dataset statistics are summarized in Tab.~\ref{tab:supp_dataset_splits}. For checkpoint selection, we randomly hold out $20\%$ of the training set as a validation split.

\begin{table*}[t]
  \centering
  \caption{Dataset splits and annotation statistics. Instance counts are measured at $1024\times1024$ resolution before cropping or quadrant partitioning.}
  \label{tab:supp_dataset_splits}
  \setlength{\tabcolsep}{4pt}
  \renewcommand{\arraystretch}{1.08}
  \begin{tabular}{@{}lllrr@{}}
    \toprule
    Dataset & Split & Source & Image pairs & Instances \\
    \midrule
    \multirow{2}{*}{BRIGHT}
      & Train & Official challenge training split & 2,332 & 186,373 \\
      & Test & Official challenge validation split & 697 & 58,579 \\
    \midrule
    \multirow{2}{*}{xBD}
      & Train & Original Tier~1 training subset & 2,799 & 159,794 \\
      & Test & Original official holdout split & 933 & 53,137 \\
    \bottomrule
  \end{tabular}
\end{table*}

\noindent\textbf{BRIGHT.} The full BRIGHT benchmark contains three damage categories~\cite{chen2025bright}.
In the public instance-level challenge release, labels are available for the
training and validation splits, while holdout labels remain private~\cite{chen2026brightchallenge}; we therefore use validation as our test set.
The official lists contain 2,332 and 697 pairs, split within disaster events.
We exclude the Ukraine, Myanmar, and Mexico events because their optical data
require separate acquisition, leaving 186,373 and 58,579 valid instances.
We additionally evaluate scene-level transfer on a separately collected
2025 Eaton Fire scene in Altadena, California.

\noindent\textbf{xBD.} From xBD~\cite{gupta2019xbd}, we use the Tier~1 training subset (2,799 pairs)
and the official 933-pair holdout split. The xView2 protocol follows a
similar-distribution, image-level split and does not enforce disjoint disaster
events, so it measures standard benchmark performance rather than strict
cross-disaster generalization. After removing \emph{un-classified}
annotations, the training and holdout splits contain 159,794 and 53,137
instances across four damage grades. We additionally evaluate large-area
transfer on a separately selected scene from the 2011 Joplin EF5 tornado.



\subsection{Damage Description Bank}
\label{sec:supp-prompts}
\label{sec:supp_prompt_design}

We construct damage descriptions based on the xBD Joint Damage Scale~\cite{gupta2019xbd} and the BRIGHT category definitions~\cite{chen2025bright}. For each damage category, we design a category-specific set of complementary descriptions to cover its diverse structural manifestations and hazard-related visual cues, including roof loss, wall collapse, structural deformation, debris, wind, fire, flooding, mud, and volcanic deposits. The number of descriptions is adapted to the visual diversity of each category, with fewer descriptions assigned to visually consistent categories such as intact/no-damage.

The complete description bank contains 19 descriptions for BRIGHT, distributed as $(3,9,7)$ across \emph{intact}, \emph{damaged}, and \emph{destroyed}, and 20 for xBD, distributed as $(3,5,5,7)$ across its four damage grades. Tab.~\ref{tab:prompts_all} lists all descriptions, and we use the complete bank for both training and inference.

\begin{table*}[t]
  \centering
  \caption{Complete damage-description bank for BRIGHT and xBD. Descriptions within each category are ordered by retention priority, from general structural evidence to hazard-specific cues.}
  \label{tab:prompts_all}
  \setlength{\tabcolsep}{5pt}
  \renewcommand{\arraystretch}{1.06}
  \begin{tabular}{@{}rl@{}}
    \toprule
    Damage Category & Description \\
    \midrule
    \multirow{3}{*}{\makecell[r]{Intact (BRIGHT) /\\ No damage (xBD)}}
      & building retaining its pre-event footprint and overall roof shape \\
      & building with continuous roof covering and no newly exposed roof sections \\
      & building with a continuous outer wall outline and no collapsed sections \\
    \midrule
    \multirow{9}{*}{\makecell[r]{Damaged\\(BRIGHT)}}
      & building with patches of roof covering missing while the roof shape is retained \\
      & building with a collapsed roof section exposing the interior while other sections remain \\
      & building with a collapsed outer wall section and part of the roof still standing \\
      & building with a tilted or shifted main roof plane and surviving standing sections \\
      & building with displaced roof-covering panels and the main roof shape still preserved \\
      & building with visible cracks in its roof or walls but no collapsed sections \\
      & building with localized burn marks and its main roof still in place \\
      & building with floodwater or mud surrounding its base while the structure remains exposed \\
      & building with volcanic material reaching its footprint while part of the structure remains exposed \\
    \midrule
    \multirow{5}{*}{\makecell[r]{Minor damage\\(xBD)}}
      & building with patches of roof covering missing while the roof shape is retained \\
      & building with displaced roof-covering panels and the main roof shape still preserved \\
      & building with visible cracks in its roof or walls but no collapsed sections \\
      & building with localized burn marks and its main roof still in place \\
      & building with floodwater around its base and its roof still exposed \\
    \midrule
    \multirow{5}{*}{\makecell[r]{Major damage\\(xBD)}}
      & building with a collapsed roof section exposing the interior while other sections remain \\
      & building with a collapsed outer wall section and part of the roof still standing \\
      & building with a tilted or shifted main roof plane and surviving standing sections \\
      & building enclosed by floodwater or mud while its roof remains visible \\
      & building with volcanic material reaching its footprint while part of the structure remains exposed \\
    \midrule
    \multirow{7}{*}{\makecell[r]{Destroyed\\(BRIGHT / xBD)}}
      & building replaced by a compact pile of collapsed roof and wall rubble \\
      & building reduced to widely scattered structural debris with no main structure standing \\
      & building burned down to a foundation with only charred wall remnants \\
      & building absent from its pre-event footprint with an exposed foundation remaining \\
      & building covered by floodwater at its pre-event location \\
      & building buried under mud and transported debris at its pre-event location \\
      & building buried beneath volcanic deposits at its pre-event location \\
    \bottomrule
  \end{tabular}
\end{table*}

\subsection{Implementation of Language-Guided Baselines}
\label{sec:supp_text_adaptations}

To compare language-integration strategies under consistent settings, we adapt six representative methods using the same bi-temporal backbone, HTC head, and text encoder. These \emph{method-style} adaptations retain each method's characteristic fusion or prompting mechanism rather than its original architecture.

\noindent\textbf{DenseCLIP-style~\cite{rao2022denseclip}.}
A two-layer Transformer decoder updates text embeddings using $8\times8$-pooled multi-scale visual features. Pixel--text similarity maps are concatenated with visual features before the FPN, with an auxiliary pixel-level BCE loss (weight $0.1$) and the standard HTC classifier.

\noindent\textbf{GLIP-style~\cite{li2022grounded}.}
Two rounds of four-head bidirectional cross-attention refine visual features and a global description bank using $8\times8$-pooled visual keys. RoIs are classified against the shared text bank through log-mean-exp aggregation, without spatial text retrieval or alignment loss.

\noindent\textbf{LAVT-style~\cite{yang2022lavt}.}
At each scale, visual tokens attend to description embeddings to generate language-conditioned features, which modulate projected visual maps through a $\tanh$-gated multiplicative residual before standard HTC classification.

\noindent\textbf{ChangeCLIP-style~\cite{dong2024changeclip}.}
A shared two-layer decoder separately contextualizes damage descriptions with pre- and post-event features. Pixel--text similarity maps, difference-guided weights, and post-event visual--text attention enhance temporal features before FPN/HTC.

\noindent\textbf{MMChange-ITFF-style~\cite{zhou2025mmchange}.}
Normalized pixel--description similarities generate spatial language features that are added to visual maps and refined through channel, spatial, and grouped pixel attention. Caption generation and text-difference modules are excluded.

\noindent\textbf{CoCoOp-style~\cite{zhou2022cocoop}.}
A two-layer meta-network converts pooled pre-event, post-event, and difference features into four learnable context tokens for the frozen BGE-small encoder. The resulting image-conditioned descriptions support log-mean-exp RoI classification without pixel--text fusion or alignment loss.

\section{Extended Experimental Results}

\subsection{Performance Across Disaster Types}

\begin{figure*}[t]
  \centering
  \includegraphics[width=\linewidth]{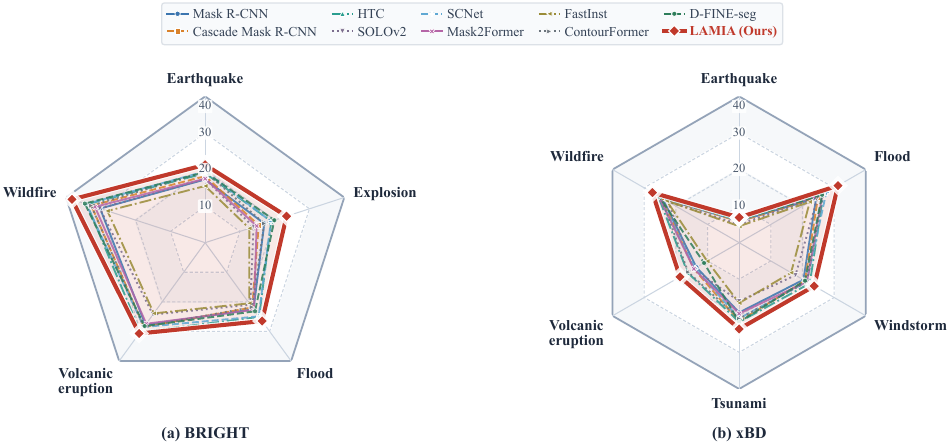}
  \caption{Mask AP comparison across disaster types on BRIGHT and xBD.}
  \label{fig:disaster_rader}
\end{figure*}

Fig.~\ref{fig:disaster_rader} compares mask AP across disaster types. On
BRIGHT, LAMIA achieves the highest AP for all five hazards. It improves over the strongest
competing method by an average of 2.2 points, with the largest gains for
wildfires and explosions. On xBD, LAMIA also leads across all six hazards. The average
improvement over the strongest competing method is 1.9 points, with the
largest gain observed for floods. These results show that the overall improvement is distributed
across multiple disaster types rather than driven by a single hazard.

\subsection{Qualitative Results and Large-Area Evaluation}

\begin{figure*}[t]
  \centering
  \includegraphics[width=\linewidth]{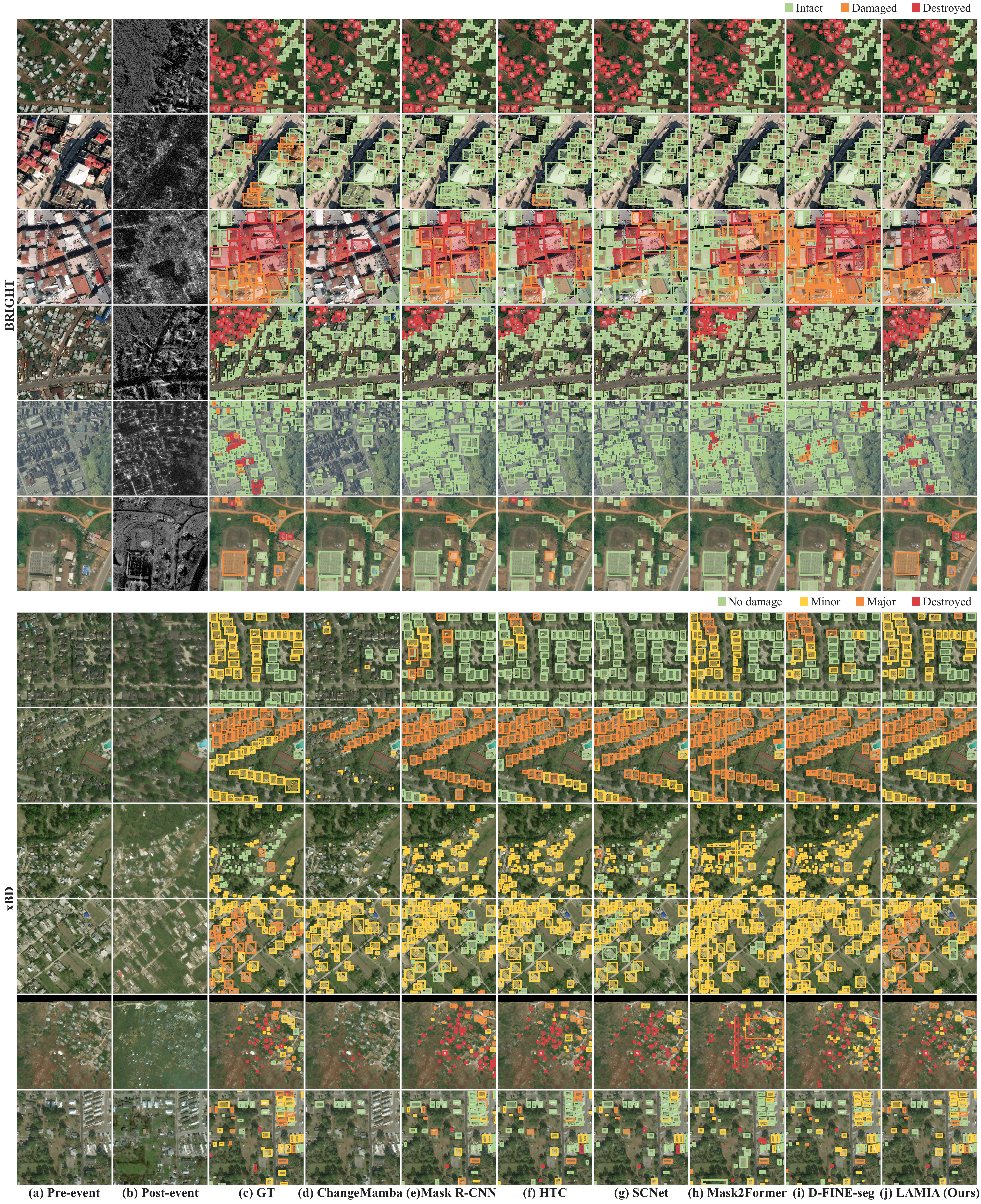}
  \caption{Additional qualitative comparisons on the BRIGHT and xBD test sets.}
  \label{fig:comparison_extend}
\end{figure*}

\begin{figure*}[t]
  \centering
  \includegraphics[width=\linewidth]{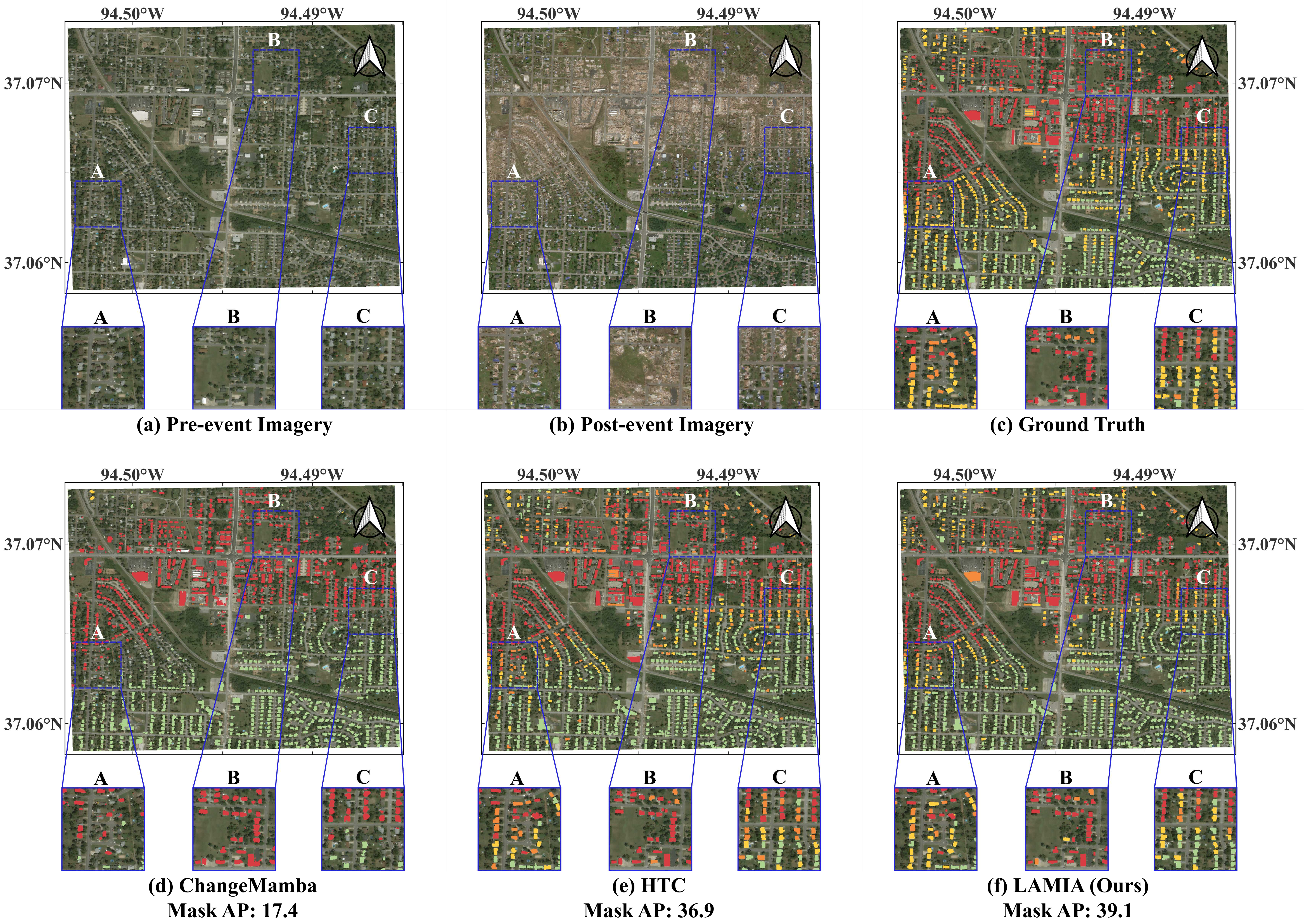}
  \caption{Large-area building damage assessment for the 2011 Joplin EF5 tornado in Missouri, USA, using pre- and post-disaster optical imagery acquired on August 8, 2009, and May 29, 2011, respectively.}
  \label{fig:joplin}
\end{figure*}

\begin{figure*}[t]
  \centering
  \includegraphics[width=\linewidth]{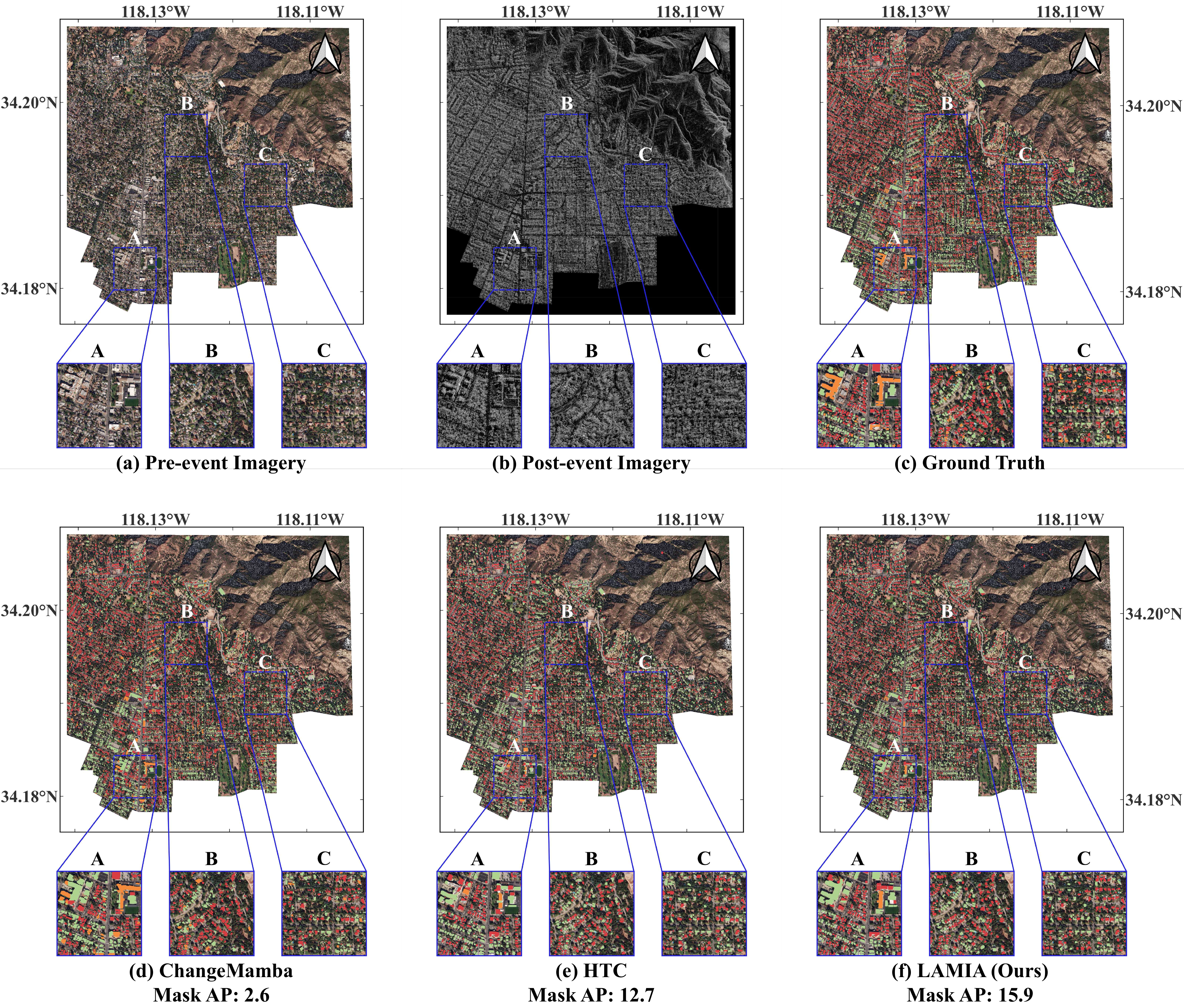}
  \caption{Large-area building damage assessment for the 2025 Eaton Fire in Altadena, California, USA, using pre-disaster optical and post-disaster SAR imagery.}
  \label{fig:california}
\end{figure*}

Fig.~\ref{fig:comparison_extend} presents additional qualitative comparisons on BRIGHT and xBD, illustrating LAMIA's improved discrimination of building damage categories. Figs.~\ref{fig:joplin} and~\ref{fig:california} further evaluate large-area damage assessment on two unseen disaster scenes: the 2011 Joplin EF5 tornado (optical--optical) and the 2025 Eaton Fire in Altadena, California (optical--SAR). LAMIA achieves mask AP scores of 39.1 on Joplin, compared with 36.9 for HTC and 17.4 for ChangeMamba. On the Eaton Fire, the corresponding scores are 15.9, 12.7, and 2.6. These results further support LAMIA's effectiveness in both optical--optical and optical--SAR settings.

\section{Further Ablation Studies}

\subsection{Semantic Adaptation and Module Coupling}

Tab.~\ref{tab:supp_grounding_final} evaluates text contextualization, instance-adaptive prototypes, and their coupling. Introducing static text prototypes reduces AP on BRIGHT from 31.5 to 30.4 and AP$_{50}$ on xBD from 55.0 to 53.6, indicating the limitations of directly matching fixed descriptions with bi-temporal visual features. Generic cross-attention provides modest improvements, reaching 32.3 and 32.0 AP on BRIGHT and xBD, respectively. ATVA alone improves AP to 33.0 on both datasets, while DIPG alone achieves 32.6 and 32.9. However, combining the two modules without shared text states reduces AP to 32.3 and 31.9, highlighting the importance of consistent semantic representations across feature fusion and instance classification. By coupling ATVA and DIPG through shared contextualized text states and alignment feedback, full LAMIA achieves the highest AP of 33.5 on BRIGHT and 33.4 on xBD, outperforming the vision-only baseline by 2.0 and 1.7 points, respectively.

\begin{table*}[t]
\centering
\caption{Ablation of text contextualization, instance-adaptive prototypes, and cross-stage coupling.}
\label{tab:supp_grounding_final}
\setlength{\tabcolsep}{3pt}
\renewcommand{\arraystretch}{1.08}
\begin{tabular}{@{}lccccc@{}}
\toprule
Variant & Text state & Prototype & Coupling & \makecell{BRIGHT\\AP/AP$_{50}$/AP$_{75}$} & \makecell{xBD\\AP/AP$_{50}$/AP$_{75}$} \\
\midrule
Vision-only baseline & -- & -- & None & 31.5/52.7/33.9 & 31.7/55.0/34.4 \\
Static text prototype & Static & Static & None & 30.4/49.7/33.0 & 31.9/53.6/34.6 \\
Generic cross-attention & Contextualized & Static & None & 32.3/53.0/35.2 & 32.0/55.0/34.5 \\
ATVA only & Contextualized & Static & Fusion only & 33.0/53.8/35.9 & 33.0/56.7/35.6 \\
DIPG only & Static & Instance-adaptive & Classification only & 32.6/52.7/35.3 & 32.9/56.5/35.6 \\
Decoupled ATVA + DIPG & Contextualized & Instance-adaptive & No shared state & 32.3/52.9/35.2 & 31.9/54.9/34.3 \\
Full LAMIA & Contextualized & Instance-adaptive & Shared state + feedback & 33.5/55.4/36.1 & 33.4/57.1/36.2 \\
\bottomrule
\end{tabular}
\end{table*}

\subsection{Visual Reliability and Description Preference}

Tab.~\ref{tab:supp_reliability_final} examines visual reliability $\rho$ and description preference $\pi$. Fixing $\rho=0.5$ reduces AP to 30.8 on BRIGHT and 31.6 on xBD, indicating that a uniform visual--language allocation is insufficient for buildings with varying visual reliability. Replacing the proposed reliability estimate with max-softmax confidence or predictive entropy improves AP to 32.4--32.5 on BRIGHT and 32.8 on xBD, but remains below full LAMIA (33.5/33.4). Batch-shuffling $\rho$ yields similar results (32.5/32.8), further supporting the importance of building-specific reliability estimation. For description preference, uniform weighting reduces AP to 31.5 and 32.0, while batch-shuffling $\pi$ achieves 32.3 and 32.8 on BRIGHT and xBD, respectively. Both alternatives underperform instance-adaptive preference estimation, demonstrating the benefit of assigning description weights according to individual building evidence.

\begin{table*}[t]
\centering
\caption{Effects of visual reliability estimation and instance-specific description preferences.}
\label{tab:supp_reliability_final}
\setlength{\tabcolsep}{3pt}
\renewcommand{\arraystretch}{1.08}
\begin{tabular}{@{}lllcc@{}}
\toprule
Variant & $\rho$ source & $\pi$ source & \makecell{BRIGHT\\AP/AP$_{50}$/AP$_{75}$} & \makecell{xBD\\AP/AP$_{50}$/AP$_{75}$} \\
\midrule
Fixed reliability & Fixed $\rho=0.5$ & Learned & 30.8/50.6/33.8 & 31.6/51.8/34.0 \\
Max-softmax reliability & Max softmax & Learned & 32.4/53.5/35.5 & 32.8/56.4/35.4 \\
Entropy-based reliability & Predictive entropy & Learned & 32.5/53.6/35.6 & 32.8/56.3/35.6 \\
Shuffled reliability & Batch-shuffled & Learned & 32.5/53.8/35.6 & 32.8/56.4/35.6 \\
Uniform preference & Auxiliary visual head & Uniform & 31.5/51.8/34.0 & 32.0/54.3/35.1 \\
Shuffled preference & Auxiliary visual head & Batch-shuffled & 32.3/53.5/35.7 & 32.8/56.5/35.5 \\
Full LAMIA & Auxiliary visual head & Instance-adaptive & 33.5/55.4/36.1 & 33.4/57.1/36.2 \\
\bottomrule
\end{tabular}
\end{table*}

\subsection{Architectural and Training Configurations}

\begin{table}[t]
  \centering
\caption{Ablation studies of text encoders, ATVA architectural settings, supervision scales, and loss weights.}
  \label{tab:summery_ablation}
  \setlength{\tabcolsep}{3pt}
  \renewcommand{\arraystretch}{1.05}
  \begin{tabular}{@{}lcc@{}}
    \toprule
    Variant & \makecell{BRIGHT\\AP/AP$_{50}$/AP$_{75}$} & \makecell{xBD\\AP/AP$_{50}$/AP$_{75}$} \\
    \midrule
    \rowcolor[HTML]{C0C0C0}
    \multicolumn{3}{@{}l}{\textit{Text encoders}} \\
    BGE & 33.5/55.4/36.1 & 33.4/57.1/36.2 \\
    CLIP & 33.3/54.0/36.3 & 33.3/56.8/36.1 \\
    Remote-CLIP & 33.3/54.4/36.1 & 33.4/54.2/36.3 \\
    \midrule
    \rowcolor[HTML]{C0C0C0}
    \multicolumn{3}{@{}l}{\textit{Architecture hyperparameters}} \\

    Window size$=4$ & 32.8/53.9/35.5 & 33.0/56.8/35.7 \\
    Window size$=8$ & 33.5/55.4/36.1 & 33.4/57.1/36.2 \\
    Window size$=16$ & 32.9/53.8/35.6 & 33.0/56.8/35.7 \\

    Attention layers$=1$ & 32.3/53.3/35.1 & 32.0/55.1/34.7 \\
    Attention layers$=2$ & 33.0/54.1/35.8 & 33.0/56.6/35.8 \\
    Attention layers$=3$ & 33.5/55.4/36.1 & 33.4/57.1/36.2 \\
    Attention layers$=4$ & 33.2/54.4/36.0 & 33.0/56.9/35.7 \\

    Text embed dims$=64$ & 30.3/49.1/29.7 & 30.6/48.0/24.4 \\
    Text embed dims$=128$ & 33.5/55.4/36.1 & 33.4/57.1/36.2 \\
    Text embed dims$=256$ & 33.1/53.8/35.0 & 32.1/46.6/31.2\\

    \midrule
    \rowcolor[HTML]{C0C0C0}
    \multicolumn{3}{@{}l}{\textit{Supervision and loss weights}} \\

    ATVA strides$=\{4\}$ & 33.1/54.2/35.8 & 33.0/56.6/35.7 \\
    ATVA strides$=\{8\}$ & 33.0/54.2/35.7 & 33.1/56.7/35.8 \\
    ATVA strides$=\{16\}$ & 33.2/54.3/36.0 & 33.0/56.7/35.6\\
    ATVA strides$=\{4,8\}$ & 33.2/54.3/36.3 & 33.0/56.5/35.7 \\
    ATVA strides$=\{8,16\}$ & 33.3/54.5/36.2 & 33.2/56.9/35.9\\
    ATVA strides$=\{4,8,16\}$ & 33.5/55.4/36.1 & 33.4/57.1/36.2 \\

    $\lambda_{\mathrm{ATVA}}=0$ & 32.9/54.3/35.8 & 32.9/56.6/35.8\\
    $\lambda_{\mathrm{ATVA}}=0.05$ & 33.2/54.5/36.1 & 33.0/56.7/35.8 \\
    $\lambda_{\mathrm{ATVA}}=0.10$ & 33.5/55.4/36.1 & 33.4/57.1/36.2 \\
    $\lambda_{\mathrm{ATVA}}=0.20$ & 33.4/54.8/36.6 & 33.0/56.6/35.8 \\

    $\lambda_{\mathrm{align}}=0$ & 33.0/54.2/36.1 & 33.0/56.5/35.8 \\
    $\lambda_{\mathrm{align}}=0.5$ & 33.3/54.7/36.4 & 33.3/56.4/36.2 \\
    $\lambda_{\mathrm{align}}=1.0$ & 33.5/55.4/36.1 & 33.4/57.1/36.2 \\
    $\lambda_{\mathrm{align}}=2.0$ & 33.3/54.7/36.5 & 33.0/56.6/35.7 \\

    \bottomrule
  \end{tabular}
\end{table}

Tab.~\ref{tab:summery_ablation} compares text encoders, ATVA configurations, supervision scales, and auxiliary loss weights under the same training protocol.

\noindent\textbf{Text encoders.}
Replacing BGE with CLIP~\cite{radford2021clip} or Remote-CLIP~\cite{liu2024remoteclip} produces only minor variations in mask AP, with differences within 0.2 points on BRIGHT and 0.1 points on xBD. This suggests that LAMIA is relatively insensitive to the choice of frozen text encoder. BGE achieves the highest AP on BRIGHT (33.5) and ties with Remote-CLIP on xBD (33.4), while obtaining the highest AP$_{50}$ on both datasets (55.4/57.1). We therefore adopt BGE as the default text encoder.

\noindent\textbf{ATVA architecture.}
A window size of $w=8$ achieves the highest AP on both datasets (33.5/33.4), outperforming $w=4$ (32.8/33.0) and $w=16$ (32.9/33.0). Increasing the number of attention layers from $L=1$ to $L=3$ consistently improves AP, whereas $L=4$ yields slightly lower performance. The shared embedding dimension also affects accuracy: $d=64$ substantially reduces AP to 30.3/30.6, while $d=256$ achieves 33.1/32.1. These results support the selected configuration of $(w,L,d)=(8,3,128)$.

\noindent\textbf{Supervision and loss weights.}
Applying ATVA supervision at all three strides $\{4,8,16\}$ achieves the highest AP (33.5/33.4), outperforming single- and two-stride configurations. Removing attention supervision ($\lambda_{\mathrm{ATVA}}=0$) reduces AP to 32.9 on both datasets, while increasing its weight to $0.20$ yields 33.4/33.0. Similarly, removing instance-level alignment ($\lambda_{\mathrm{align}}=0$) reduces AP to 33.0 on both datasets, and increasing its weight to $2.0$ yields 33.3/33.0. The selected weights, $\lambda_{\mathrm{ATVA}}=0.10$ and $\lambda_{\mathrm{align}}=1.0$, achieve the highest overall AP on both benchmarks.

\section{Limitations}
\label{sec:supp_limitations}

LAMIA primarily focuses on improving damage-grade classification through language-guided semantic alignment, rather than explicitly refining building localization or mask boundaries. In addition, it employs a shared description bank across disaster types without hazard-specific adaptation. Incorporating disaster-aware semantic descriptions and conditioning mechanisms may further improve its applicability to diverse hazards.


\end{document}